\documentclass{article} 
\usepackage{iclr2027_conference,times}

\usepackage{amsmath,amsfonts,bm}

\def\eqref#1{equation~\ref{#1}}

\def\1{\bm{1}}

\def\vc{{\bm{c}}}

\def\vk{{\bm{k}}}

\def\vq{{\bm{q}}}

\def\mI{{\bm{I}}}

\def\mK{{\bm{K}}}

\def\mSigma{{\bm{\Sigma}}}

\DeclareMathAlphabet{\mathsfit}{\encodingdefault}{\sfdefault}{m}{sl}
\SetMathAlphabet{\mathsfit}{bold}{\encodingdefault}{\sfdefault}{bx}{n}

\newcommand{\E}{\mathbb{E}}

\newcommand{\Var}{\mathrm{Var}}

\newcommand{\Cov}{\mathrm{Cov}}

\newcommand{\mymethod}{TesseraKV}

\newcommand{\armquest}{\textsc{Quest}}

\newcommand{\dhead}{{d}}          
\newcommand{\nkeys}{{n}}          
\newcommand{\npage}{{M}}          
\newcommand{\csize}{{c}}          
\newcommand{\gqa}{{G}}            

\newcommand{\query}{\vq}                       
\newcommand{\key}[1]{{\vk_{#1}}}               
\newcommand{\cent}[1]{{\vc_{#1}}}              
\newcommand{\keyw}[1]{{\widetilde{\vk}_{#1}}}  
\newcommand{\centw}[1]{{\widetilde{\vc}_{#1}}} 

\newcommand{\Sq}{{\mSigma_\mathrm{q}}}         
\newcommand{\Sstar}{{\mSigma^{\star}}}        
\newcommand{\hSig}{{\widehat{\mSigma}}}       
\newcommand{\Sqhalf}{{\Sq^{1/2}}}

\newcommand{\logit}[1]{{\ell_{#1}}}            
\newcommand{\pmap}{{\pi}}                      
\newcommand{\gscore}[1]{{\gamma_{#1}}}         
\newcommand{\wset}{{\mathcal{W}}}              
\newcommand{\clus}[1]{{\mathcal{P}_{#1}}}      
\newcommand{\phimap}{{\phi}}                   

\newcommand{\museg}{Muse-Glimmer-30B}
\newcommand{\northm}{North-Mini-Code-30B}
\newcommand{\cmdr}{Command-R-7B}

\newcommand{\Sctx}{{\mSigma_\mathrm{ctx}}}      
\newcommand{\Sqst}{{\mSigma_\mathrm{qst}}}      
\newcommand{\Sbar}{{\overline{\mSigma}}}        
\newcommand{\spread}[1]{{s_{#1}}}               
\newcommand{\mixw}{{\lambda}}                   
\newcommand{\kgate}{{\kappa}}                   
\newcommand{\budget}{{B}}                       
\newcommand{\nsink}{{n_\mathrm{s}}}             
\newcommand{\Rset}{{\mathcal{R}}}               
\newcommand{\Cset}{{\mathcal{C}}}               
\newcommand{\Oset}{{\mathcal{O}}}               
\newcommand{\Sel}{{\mathcal{S}}}                

\usepackage[utf8]{inputenc} 
\usepackage[T1]{fontenc}    
\usepackage{hyperref}       
\usepackage{url}            
\usepackage{booktabs}       
\usepackage{amsfonts}       
\usepackage{nicefrac}       
\usepackage{microtype}      
\usepackage{xcolor}         
\usepackage{multirow}

\usepackage{xcolor}
\usepackage{graphicx}
\usepackage[capitalize,noabbrev]{cleveref}
\usepackage{caption}
\usepackage{subcaption}
\usepackage{wrapfig}
\usepackage{amsthm}
\usepackage{makecell}
\usepackage{float}
\usepackage{soul}
\usepackage[normalem]{ulem} 

\usepackage{algorithm}
\usepackage{algpseudocode}
\algrenewcommand{\algorithmicrequire}{\textbf{Input:}}
\algrenewcommand{\algorithmicensure}{\textbf{Output:}}

\definecolor{qkred}{HTML}{C5172E}
\definecolor{vogreen}{HTML}{118B50} 
\definecolor{ffnblue}{HTML}{1B56FD}

\newcommand{\reviewOne}[1]{\textcolor{green}{@Reviewer YCYE}}
\newcommand{\reviewTwo}[1]{\textcolor{orange}{@Reviewer UPhA}}
\newcommand{\reviewThree}[1]{\textcolor{purple}{@Reviewer 5D7g}}

\crefname{problem}{Problem}{Problems}
\Crefname{problem}{Problem}{Problems}
\crefname{assumption}{Assumption}{Assumptions}
\Crefname{assumption}{Assumption}{Assumptions}
\crefname{definition}{Definition}{Definitions}
\Crefname{definition}{Definition}{Definitions}
\crefname{remark}{Remark}{Remarks}
\Crefname{remark}{Remark}{Remarks}

\title{ 
\mymethod{}: KV Paging via Co-Retrieval
}

\author{
Jeffrey T. H. Wong\thanks{Part of this work was carried out during an internship at MediaTek Research.}$^{*1}$
, Taowen Liu$^{1}$, Can Xiao$^{1}$, Cheng Zhang$^{1}$,\\ {\bf Qian Zuo$^{1}$, Bryan Tan$^{1}$, Davide Buffelli$^{2}$, Alberto Bernacchia$^{2}$, Yiren Zhao$^{1}$} \\
\\
$^{1}$Imperial College London \quad $^{2}$ MediaTek Research \quad $^{3}$ University of Edinburgh \\
\texttt{\{tsz.wong20, tony.liu20, can.xiao22\}@imperial.ac.uk} \\ 
\texttt{\{cheng.zhang122, bryan.tan20, a.zhao\}@imperial.ac.uk} \\ 
\texttt{qian.zuo@ed.ac.uk} \quad 
\texttt{\{davide.buffelli, alberto.bernacchia\}@mtkresearch.com}
}

\iclrfinalcopy 
\begin{document}

\maketitle

\begin{abstract}
Agentic workloads increasingly operate over long contexts, causing KV caches to grow rapidly and making memory bandwidth a major bottleneck during generation. KV paging
\footnote{We use \emph{KV paging} to denote the partitioning of the KV cache into pages that serve as units of retrieval. This term refers to how keys are organized and scored for retrieval, and is different from memory-management mechanisms like PagedAttention, which organize the KV cache into blocks for efficient memory allocation.}
alleviates this cost by retaining the cache while retrieving only the pages needed at each decoding step. 
There are two critical challenges in KV paging: how keys are grouped into pages
and how pages are scored for retrieval. Existing methods construct
pages using heuristics, and score those pages using a separate criterion. 
These two decisions should instead be derived from a common objective: 
whether the keys in a page are likely to be retrieved together.
We propose \mymethod{}, which derives both page construction and page scoring
from \emph{co-retrieval}: keys should share a page when their attention logits
tend to vary together under the query distribution of a head. This common
principle yields a covariance-weighted geometry for partitioning keys and a
spread-aware score for estimating the attention mass of each page. Because the
decoding-query distribution is unavailable when the cache is constructed, we
further study how to estimate this geometry from queries available beforehand.
We find that prefill queries provide a useful but task-dependent signal for
decoding-query geometry, while incorporating queries from the downstream
instruction substantially improves the co-retrieval structure relevant to
partitioning.
Across a range of LLMs with NoPE, long-context tasks, and agentic workloads, \mymethod{} consistently improves downstream generation quality over state-of-the-art KV retrieval methods at matched KV budgets.
For example, on \textsc{Claw-Eval} agentic tasks, \mymethod{} achieves an overall success rate of $70.32$, improving over the previous state of the art at $42.58$ by $27.74$ points. We further report empirical results of \mymethod{} on RoPE models and its integration with KV CPU offloading.

\end{abstract}

\section{Introduction}
\label{sec:intro}

\begin{figure}[t]
    \centering
    \includegraphics[width=\linewidth]{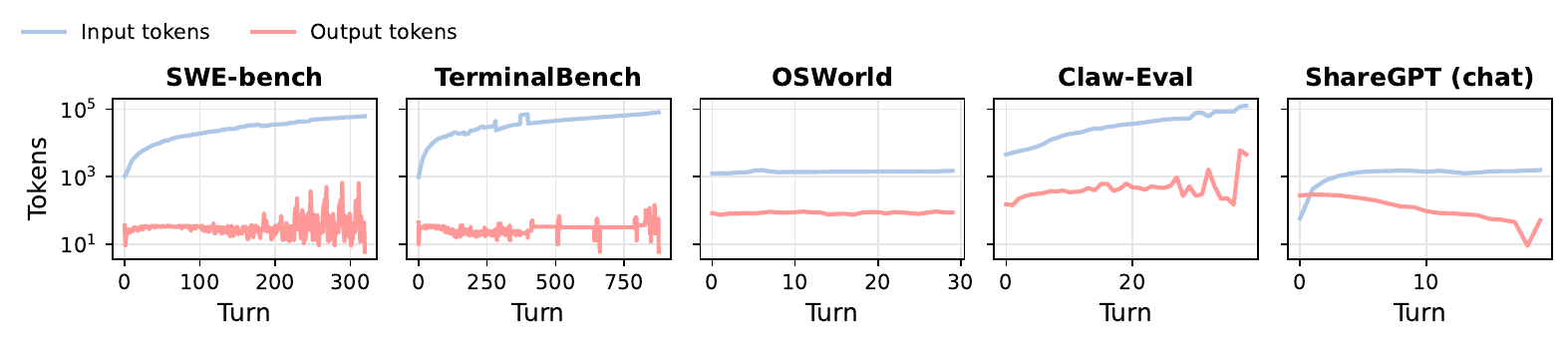}
    \vspace{-2em}

    \caption{
    Input and output tokens per turn across four agentic workloads and open-ended chat. Lines show median tokens at each turn over active sessions. Agentic workloads accumulate input tokens faster across turns while output token counts remain relatively stable, in contrast to open-ended chat.
    }

    \vspace{-1em}
    \label{fig:01:motivation}
\end{figure}

Agentic workloads have become a central use of large language models. 
Unlike traditional chat interaction, where each turn contributes comparable numbers of prompt and response tokens, 
agentic workloads accumulate context through tool outputs, file contents, and search results far exceeding token generation.
Across turns, the consumed context therefore grows
steadily while the number of generated tokens remains relatively flat. 
As a result, the KV cache expands rapidly, and reading from it increasingly
dominates the decoding latency. 
\Cref{fig:01:motivation} shows that input and
output token counts follow distinctly different trends across turns in
agentic workloads and chat interaction. 
Over a \textsc{SWE-bench}~\citep{swe-bench} trajectory, the median context
read grows by a factor of $100$ from the first to the last turn. Summed over a
session, an agent reads more than $200$ context tokens for every token it
generates, compared with $2.3\times$ for open-ended chat~\citep{sharegpt}.
Similar trends are observed in \textsc{TerminalBench}~\citep{terminalbench}
and \textsc{Claw-Eval}~\citep{claw-eval}.

To address this challenge, recent studies have proposed methods for computing
sparse attention. One line of work uses \emph{eviction}: tokens scored as
unimportant are discarded from the KV cache
\citep{snapkv, zhang2023h2o, park2026keydiff}. Once discarded, these
tokens cannot be recalled in the future. In a long trajectory, this can be
dangerous: a token that appears irrelevant at one point may become crucial for
a later turn. 
An alternative is \emph{retrieval}, which preserves the full KV cache while selectively accessing only the portions relevant to the current query
\citep{quest, clusterkv, oasisKV, chen2025retroinfer}.
Since scoring every key individually would
defeat the purpose of retrieval, keys are grouped into pages that are fetched as a
unit. We use the term \emph{KV paging} to emphasize the two coupled decisions in this process:
\emph{partition}, which determines how the keys are grouped, and \emph{gate}, which determines how those pages are scored for retrieval.

Existing methods make these two decisions based on heuristics and treat them independently. Position-based approaches group contiguous tokens \citep{quest, arkvale}, assuming that attention is local; similarity-based approaches group keys with high cosine similarity \citep{clusterkv, louiskv}, assuming that semantically close keys are retrieved together.
These approaches characterize key relationships independently of the queries that determine retrieval. 
In contrast, we characterize relationships between keys through how their attention logits co-vary under the query distribution, directly capturing their tendency to be retrieved together.
%

We refer to this tendency as \emph{co-retrieval} and use it to jointly design the partition and gate. We show that 
co-retrieval induces a covariance-weighted geometry over key vectors: the resulting partition minimizes co-retrieval spread within each page, while the gate accounts for the residual spread that no finite partition can eliminate.
We highlight the following contributions of \mymethod{}:

\vspace{-0.5em}
\begin{itemize}
\item We introduce \emph{co-retrieval}, a novel objective that directly optimizes whether keys can be retrieved together under a given query distribution under NoPE attention. From this objective, we derive an efficient $k$-means formulation for page partitioning and a closed-form solution for the gate.

\item We design metrics and estimators for predicting the query distribution at decode-time. We propose a simple \emph{query-mix} estimator that combines queries from the context with queries generated from the downstream instruction, effectively closing the gap to the reference on the co-retrieval metric.

\item We conduct extensive experiments across a range of NoPE LLMs and show
that \mymethod{} substantially outperforms SoTA 
methods. \mymethod{} achieves an overall success rate of $70.32$ on \textsc{Claw-Eval}, improving over the previous state of the art at $42.58$ by $27.74$ points. We  provide empirical results on its application on RoPE models.
\end{itemize}

\section{Background and Related Work}
\label{sec:relatedwork}

\paragraph{Long-Context Models}
Context windows have grown from a few thousand tokens to millions of tokens.
Recent long-context models increasingly adopt hybrid architectures
\citep{glm-5.3-flash, qwen3.8flashnext, kimik3, museglimmer30b,
north-mini-code, cmdr-7b}, 
interleaving local layers based on either sliding-window attention or gated linear recurrence \citep{gated-delta-networks} with a small number of global attention layers
that enable long-range understanding and retrieval. These global layers are often free
of positional encoding \citep{rope-to-nope, DroPE}. Consequently, the cost of
serving long inputs is concentrated in these layers. Their dominant cost is
the KV cache, which has shape $\text{seq} \times \text{heads} \times \dhead$
per layer and must be read in full at every decoding step, making memory
bandwidth the primary bottleneck to generation. Architectural design reduces two of the three dimensions of the KV cache.
Sharing KV representations across query heads \citep{mqa, gqa} reduces the head dimension.
The feature dimension can be reduced either architecturally, such as MLA and low-rank approaches~\citep{mla, a3, svd-llm},
or by lowering the precision of the KV elements stored in the cache~\citep{Turboquant, KVarN}.

\paragraph{KV Cache Eviction}
Eviction methods compress the sequence axis at inference time by permanently discarding low-scoring KV entries. 
Attention-based methods score keys from the attention matrix \citep{snapkv, zhang2023h2o, cai2024pyramidkv, agentkv},
while key-based methods use geometric criteria such as distance from local centroids or from the key manifold \citep{park2026keydiff, manifoldkv}. 
\emph{Expected Attention}~\citep{ea} is most closely related to our work: it models head queries as Gaussian and scores each key by its expected attention under that distribution.
Eviction leaves the decoding path unchanged but is irreversible. A dropped key required by a later query is unavailable rather than merely deprioritised, and performance degrades at tight budgets.

\paragraph{KV Cache Retrieval}
Retrieval methods avoid permanent eviction by retaining the KV cache and grouping keys into \emph{pages} that are fetched as a unit.
The cache may be offloaded to CPU memory or another memory tier.
At each decoding step, they select only the pages relevant to the current query, leaving the remaining cache available for future steps. Hence this introduces two design decisions in the choice of the \emph{partition} and the \emph{gate}.
\armquest{}~\citep{quest} and ArkVale~\citep{arkvale} partition keys by position
and score a page using an upper bound on the largest logit it may contain.
ClusterKV~\citep{clusterkv} and LouisKV~\citep{louiskv}, in contrast, partition
keys by content using cosine similarity and score each page from its centroid.
RetroInfer~\citep{chen2025retroinfer} takes a complementary approach: it estimates the attention contribution of clusters that are not retrieved, instead of treating their contribution as zero.

Both partitioning strategies rely on a heuristic assumption: tokens that are similar in key space or position are likely to be retrieved together. We instead characterize co-retrieval through how attention logits vary together under each head's query distribution under the NoPE attention, and use this relationship to derive both the partition objective and the page-scoring rule. 

\begin{figure}[t]
    \centering
    \includegraphics[width=\linewidth]{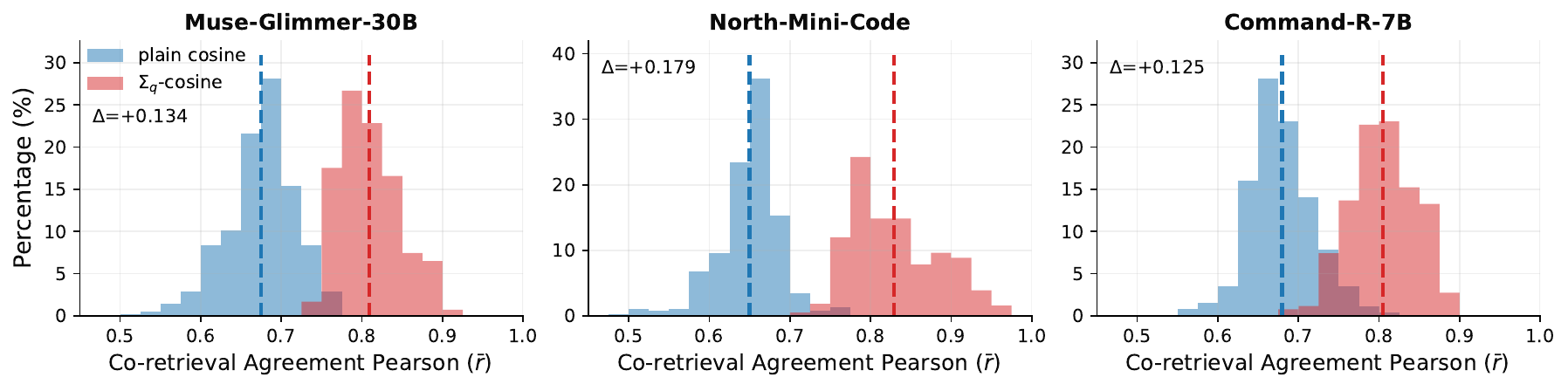}
    \vspace{-2em}
    \caption{Co-retrieval agreement between predicted  and empirical
    attention-logit correlations for three models.
    Each panel
    shows the distribution of per-head Pearson correlations $\bar r$ across
    attention heads, averaged over six domains from SlimPajama~\citep{slimpajama}. 
    Distributions concentrated further to the right indicate higher Pearson correlations and thus more accurate prediction of co-retrieval.
    }
    \vspace{-1em}
    \label{fig:03:prefill}
\end{figure}

\section{\mymethod}
\label{sec:coretrieval}

We first formalize this co-retrieval principle by modeling queries as a distribution in~\Cref{sec:key-retrieval}, then derive the partition solution in~\Cref{sec:partition} and a corresponding gating strategy in~\Cref{sec:spread-gate}. Since the distribution of queries at decode time is unavailable when pages are constructed, we discuss how to estimate its covariance from the context and downstream instruction in~\Cref{sec:cov-estimation}.

\subsection{Key Retrieval in Autoregressive Transformers}
\label{sec:key-retrieval}

We consider a single global attention head without positional encoding (NoPE) in a decoder-only transformer~\citep{attention-is-all-you-need}, with head dimension $\dhead$.
At decoding step $t$, the query $\query_t \in \mathbb{R}^{\dhead}$ scores each cached key $\key{i} \in \mathbb{R}^{\dhead}$ through
\begin{equation}
  a_{ti}
  = \frac{\exp(\logit{ti})}{\sum_{j \le t} \exp(\logit{tj})} ,
  \qquad
  \logit{ti} = \frac{\query_t^{\top} \key{i}}{\sqrt{\dhead}} ,
  \label{eq:attention}
\end{equation}
where $\logit{ti}$ is the attention logit and $a_{ti}$ is the normalized attention weight.

\paragraph{Co-retrieval.}
We say that a query $\query_t$ \emph{retrieves} a key $\key{i}$ when the attention score
$a_{ti}$ is large. Retrieval is therefore a property of the pair $(\query_t, \key{i})$ instead of the key alone, and a page is well formed when its members are
retrieved by the same queries.  
Since we construct
the pages independently of the any particular query at inference time, it is useful to
characterize co-retrieval through a query distribution.

We therefore model the queries arriving at a given head as draws of a random vector $\query$ with covariance
$\Sq = \Cov(\query) \in \mathbb{R}^{\dhead \times \dhead}$, and defer the question of how $\Sq$
is estimated to Section~\ref{sec:cov-estimation}. 

Writing $\logit{i}(\query) = \query^{\top} \key{i} / \sqrt{\dhead}$
for the logit of key $\key{i}$ under such a query, we say that two keys are \emph{co-retrieved} to the
extent that their logits fluctuate together:
\begin{equation}
  \rho_{ij}
  \;=\; \frac{\Cov\big(\logit{i}(\query), \logit{j}(\query)\big)}
             {\sqrt{\Var\big(\logit{i}(\query)\big)\,\Var\big(\logit{j}(\query)\big)}}
  \;=\; \frac{\key{i}^{\top} \Sq \key{j}}
             {\sqrt{\key{i}^{\top} \Sq \key{i}}\;\sqrt{\key{j}^{\top} \Sq \key{j}}}
  \;=\; \cos\big(\keyw{i}, \keyw{j}\big) ,
  \qquad \keyw{i} := \Sqhalf \key{i} .
  \label{eq:corr}
\end{equation}

where the expectations are taken over $\query$. The co-retrieval correlation is thus a cosine similarity with covariance-weighted keys,
which measures angles in the geometry induced by $\Sq$. Setting $\Sq = \mI$
recovers the plain cosine similarity adopted in prior work \citep{clusterkv, louiskv}, grouping
keys by cosine similarity. \textit{When queries are not isotropic, the two measures can disagree in their ordering: a pair of keys with high cosine similarity in the raw
space may be co-retrieved far less often than a pair that appears distant, and
grouping by raw similarity therefore misplaces keys}.

To validate this, we estimate an empirical covariance $\widehat{\Sq}$ from
the prefill queries of the first 16k tokens of each SlimPajama document, and
measure how well each similarity predicts the logit correlation observed on a
held-out set of queries. \cref{fig:03:prefill} shows that the
covariance-weighted similarity is substantially more accurate than the plain
cosine similarity (see Appendix~\ref{app:coretrieval} for details).

While the correlation captures whether two keys tend to receive high or low logits together, it discards the magnitude of these fluctuations. To derive a partition objective, we therefore also consider the corresponding logit covariance.
\begin{equation}
\dhead\,\Cov\big(\logit{i}(\query),\logit{j}(\query)\big)
=\key{i}^{\top}\Sq\key{j}
\;=\; \langle\keyw{i},\keyw{j}\rangle.
\label{eq:logit-cov}
\end{equation}
Appendix~\ref{app:coretrieval-cov} evaluates the agreement between predicted and empirical logit covariances.
We next derive a partition objective from these logit relationships.

\subsection{Partitioning the Cache by Co-Retrieval}
\label{sec:partition}

\paragraph{Partition.}
We partition the KV cache into $\npage$ pages $\mathcal P = \{\clus{1}, \ldots, \clus{\npage}\}$, each 
tagged
by its key centroid $\cent{p} = |\clus{p}|^{-1}\sum_{i\in\clus{p}}\key{i}$.
At each decoding step, attention is restricted to the pages with the highest centroid scores under a fixed KV-cache budget, while unselected pages remain resident in HBM but are not accessed.

\paragraph{Partition objective.}
Because a page is retrieved as a whole, its members should be retrieved together.
We therefore seek a partition in which each member's logit fluctuates together with its page centroid's logit, minimizing the variance of their difference across queries.
For
$\key{i}\in\clus{p}$, write its logit as
\begin{equation}
  \logit{ti}
  = \underbrace{\frac{\query_t^T \cent{p}}{\sqrt{\dhead}}}_{g_{tp}}
  + \underbrace{\frac{\query_t^T (\key{i} - \cent{p})}{\sqrt{\dhead}}}_{\delta_{ti}} ,
  \label{eq:decomp}
\end{equation}
  where $g_{tp}$ is the page score and $\delta_{ti}$ is the member's deviation from it.
  Under the query model in \cref{eq:corr}, the deviation has variance
  \begin{equation}
  \Var_t(\delta_{ti})
  = \frac{(\key{i}-\cent{p})^T\Sq(\key{i}-\cent{p})}{\dhead}
  = \frac{\lVert\keyw{i}-\centw{p}\rVert^2}{\dhead},
  \label{eq:scatter}
  \end{equation}
  where $\keyw{i}=\Sqhalf\key{i}$ and
  $\centw{p}=\Sqhalf\cent{p}$.
  Since the transformation is linear, $\centw{p}$ is the mean of the transformed keys in page $p$.
  Summing the variances over all pages gives the partition objective
  \begin{equation}
  J(\mathcal P)
  := \dhead\sum_{p=1}^{\npage}\sum_{i\in\clus{p}}\Var_t(\delta_{ti})
  = \sum_{p=1}^{\npage}\sum_{i\in\clus{p}}
  \lVert\keyw{i}-\centw{p}\rVert^2.
  \label{eq:objective}
  \end{equation}
We therefore apply \textbf{Euclidean $k$-means} to the transformed keys $\keyw{i}$ to minimize this objective.
The objective also admits a pairwise interpretation,
\begin{equation}
\sum_{i\in \clus{p}} \lVert \keyw{i} - \centw{p}\rVert^2
= \frac{1}{2|\clus{p}|}\sum_{i,j\in \clus{p}} \lVert \keyw{i} - \keyw{j}\rVert^2
= \frac{\dhead}{2|\clus{p}|}\sum_{i,j\in \clus{p}}
\Var_t\big(\logit{ti} - \logit{tj}\big) ,
\label{eq:pairwise}
\end{equation}
so minimizing the objective reduces the variance of logit differences between keys in the same page, promoting co-retrieval.
We analyze its overhead in~\Cref{fig:latency}.

\paragraph{Logit correlation versus covariance.}
\label{methods:cov_vs_cor}
Expanding the pairwise variance in~\cref{eq:pairwise} gives $
\dhead\,\Var_t(\logit{ti}-\logit{tj})
= \sigma_i^2 + \sigma_j^2 - 2\sigma_i\sigma_j\rho_{ij}$,
where $\sigma_i^2 = \key{i}^T\Sq\key{i}$.
Thus, the objective accounts for both logit correlation and fluctuation scale.
Partition optimizing logit correlation can be obtained using \textbf{Spherical $k$-means} on normalized transformed keys.
In practice, we observe optimizing logit covariance performs better empirically and is our default. Appendix~\ref{app:gate-form} details the ablation.

Although the partition minimizes within-page spread, a finite number of pages generally leaves nonzero variation.
We next derive a centroid-score correction that accounts for this residual spread.

\subsection{Spread-Corrected Page Scoring}
\label{sec:spread-gate}
The centroid logit, $g_{tp} = \frac{\query_t^T\cent{p}}{\sqrt{\dhead}}$, provides a natural score for page $p$.
However, the centroid does not capture the variation of the individual keys within the page. 
To quantify the effect of within-page variation on page retrieval, we consider the average unnormalized attention mass of page $p$,

\begin{equation}
U_t(p)
:= {\frac{1}{|\clus{p}|}}  \sum_{i\in\clus{p}} 
\exp(\logit{ti})
= \exp(g_{tp}){\frac{1}{|\clus{p}|}}  \sum_{i\in\clus{p}} \exp(\delta_{ti}),
\label{eq:page-mass}
\end{equation}
The centroid is the mean of the keys,
 $\sum_{i\in\clus{p}}\delta_{ti} = 0$ for all possible query. Applying Jensen's inequality:
\begin{equation}
{\frac{1}{|\clus{p}|}} \sum_{i\in\clus{p}} \exp({\delta_{ti}})
\geq
\exp\left({\frac{1}{|\clus{p}|}\sum_{i\in\clus{p}}\delta_{ti}}\right) = 1
\end{equation}
If we estimate a page's mass using only the
centroid, this mass would be
$\exp(g_{tp})$, which is a smaller estimation than the real $U_t(p)$.
To estimate this factor without scoring individual keys, we model each
member's deviation across the query distribution as
$\delta_{ti}\sim\mathcal N(0,\spread{i}^2)$, where
$\spread{i}^2=\Var_t(\delta_{ti})$. Then
\begin{equation}
  \E_t\!\left[ 
  \exp(\delta_{ti})
  \right] = \exp\!\left(\frac{\spread{i}^2}{2}\right),
  \qquad
  \spread{i}^2 = \frac{(\key{i} - \cent{p})^T \Sq\, (\key{i} - \cent{p})}{\dhead} .
  \label{eq:mgf}
\end{equation}
Approximating the individual spreads by their page mean,
\begin{equation}
\spread{p}^2
= \frac{1}{|\clus{p}|}\sum_{i\in\clus{p}}\spread{i}^2
= \frac{1}{|\clus{p}|}\sum_{i\in\clus{p}}
\frac{\lVert\keyw{i}-\centw{p}\rVert^2}{\dhead},
\label{eq:spread}
\end{equation}
and substituting back to $U_t(p)$ gives
\begin{equation}
\log U_t(p)
\approx
g_{tp}
+\frac{\spread{p}^2}{2}.
\end{equation}
This gives the spread-corrected gate
\begin{equation}
\operatorname{score}_t(p)
=\frac{\query_t^T\cent{p}}{\sqrt{\dhead}}
+\frac{\spread{p}^2}{2}.
\label{eq:spread-gate}
\end{equation}

The spread correction complements the partition objective: 
$k$-means minimizes within-page spread, while the gate accounts for 
residual spread that remains under a finite page budget. 
The spread is computed from the resulting partition with minimal overhead.
In practice, we use a linear bias, $\kappa'\spread{p}$, which is less sensitive to coefficient choice than the quadratic form.
See Appendix~\ref{app:gate-form} for details.

\subsection{Estimating the Query Covariance}
\label{sec:cov-estimation}
The partition must be fixed before the decoding queries are
generated, hence $\Sq$ must be estimated from queries available beforehand.
Prior work models the queries of a head as Gaussian and estimates its moments from
the prefill queries of the same head
\citep{ea,training-free-activation-sparsity}. 
\Cref{fig:03:prefill} shows that covariance estimated from prefill queries predicts the co-retrieval structure of held-out decoding queries.

In a long-context downstream task, however, the prefill and decoding queries
need not come from the same distribution. The prefill queries are dominated by
the document, whereas the decoding queries are generated in response to the
downstream instruction. Their covariances can therefore differ, in which case
the prefill estimate induces the wrong geometry for both the partition and the
gate. This section quantifies the resulting gap and the extent to which it can
be closed.

\paragraph{Metrics.}
We separate two questions: how well an estimator $\hSig$ approximates the
covariance of the decoding queries characterized by \emph{covariance agreement}
, and how much its error affects \emph{co-retrieval agreement}
. Let $\Sstar$ denote the \emph{reference} 
covariance, computed from
the decoding queries themselves. 
\emph{covariance agreement}
\begin{equation}
  \mathcal A(\hSig) \;=\;
  \frac{\langle \hSig, \Sstar\rangle_F}
       {\lVert \hSig\rVert_F\,\lVert \Sstar\rVert_F},
  \label{eq:cov-agreement}
\end{equation}
is the cosine similarity between the two covariance matrices under the
Frobenius inner product. This metric is invariant to overall scale, as is the
resulting partition.

To compute the \emph{co-retrieval agreement}, for each head, we use
$\hSig$ in place of $\Sq$ in \cref{eq:corr} to obtain predicted correlations
$\hat\rho_{ij}$. We then compute the empirical correlations of the logits
$\logit{ti}$ and $\logit{tj}$ across the decoding queries and report the
Pearson correlation between the predicted and empirical correlations over
all key pairs. This measures how well the estimated covariance captures the
pairwise logit relationships on which co-retrieval depends. An estimator can
therefore have high covariance agreement yet low co-retrieval agreement if
its error is concentrated in directions along which the keys vary, and
vice versa.

\paragraph{Estimators.} 
We compare two estimators against $\Sstar$.
\emph{Prefill} follows prior work and uses the sample covariance of all
prefill queries for the head, centered by their mean.
\emph{Query-mix} separates queries produced by the document from those
produced by the question or downstream instruction that follows it, and forms the convex
combination
\begin{equation}
  \hSig_\mixw \;=\; (1-\mixw)\,\Sctx + \mixw\,\Sqst,
  \qquad \mixw\in[0,1],
  \label{eq:question-mix}
\end{equation}
Before mixing, we rescale the question covariance to match the trace of the context covariance, so that $\mixw$ controls the relative contribution of the two covariance structures. We evaluate them in~\Cref{sec:exp:estimation-acc}.

\paragraph{GQA.}
\label{sec:gqa}
Under multi-head attention, each query head has its own keys, so the
partition is constructed independently for each head using its covariance
$\Sq$. Grouped query attention instead shares a single set of keys among $\gqa$ query heads, each with its own covariance $\Sq^{(g)}$. The group therefore also needs to share a single partition to preserve bandwidth savings.

For each query head, the covariance induces a distance between keys through
the transformed-key geometry of \cref{eq:logit-cov}. We combine these head-specific
geometries by averaging their pairwise squared distances, which is equivalent
to using the average covariance. 
\begin{equation}
  (\key{i} - \key{j})^{\top} \Sbar \, (\key{i} - \key{j})
  \;=\; \frac{1}{\gqa} \sum_{g=1}^{\gqa}
        (\key{i} - \key{j})^{\top} \Sq^{(g)} \, (\key{i} - \key{j}) ,
  \qquad \Sbar = \frac{1}{\gqa}\sum_{g=1}^{\gqa} \Sq^{(g)} ,
  \label{eq:avg-graph}
\end{equation}
We therefore construct a single partition by running $k$-means under
$\Sbar$, clustering keys according to the geometry shared across the
query heads. \Cref{alg:prefill,alg:decode} state the procedure. 
We assess the accuracy cost of sharing a partition across query heads in Appendix~\ref{app:gqa}.

\section{Experiments}
\label{sec:exp}

\subsection{Setup}
\label{sec:exp:setup}

\paragraph{Baselines}
We compare \mymethod{} against two families of KV-cache methods.
Eviction methods discard keys during pre-fill and cannot recover them,
including 
SnapKV~\citep{snapkv}, KeyDiff~\citep{park2026keydiff}  and AgentKV~\citep{agentkv}
.
Retrieval methods retain the full cache and select a subset at each decoding step,
including Quest~\citep{quest}, ArkVale~\citep{arkvale}
and LouisKV~\citep{louiskv}.
All baselines are run at the same key budget and the same average page size $\csize = 64$. 
We additionally compare with RetroInfer~\citep{chen2025retroinfer}, which estimates the contribution of non-retrieved clusters instead of omitting it, with the KV cache held both in GPU memory and in CPU memory in Appendix~\ref{app:retroinfer}.
More details about the setup are available in Appendix~\ref{app:exp}.

\begin{table}[!t]
\caption{
\mymethod{} outperforms majority of the baselines on \textsc{ruler} with 8K and 16K context length. We show average score with decreasing KV budget across baselines. FullCache denotes attention over the entire KV cache. Best results for each compression ratio are displayed in bold.
}

  \label{tab:main16k}
  \centering
  \small
  \resizebox{\textwidth}{!}{%

    \begin{tabular}{@{}llcccccc|cccccc@{}}
      \toprule
      \multirow{2}{*}{Model} & 
      &
      \multicolumn{6}{c}{\textsc{ruler}-8K} & \multicolumn{6}{c}{\textsc{ruler}-16K} \\
      \cmidrule(lr){3-8} \cmidrule(l){9-14}
      & KV budget $\budget$ & $16$ & $32$ & $64$ & $128$ & $256$ & Avg. & $16$ & $32$ & $64$ & $128$ & $256$ & Avg. \\ \midrule
      \multirow{6}{*}{\museg{}} & FullCache & \multicolumn{5}{c}{94.61} & 94.61 & \multicolumn{5}{c}{94.52} & 94.52 \\ \cmidrule(l){2-14}
       & KeyDiff & 17.50 & 22.32 & 27.34 & 34.41 & 43.82 & 29.08 & 16.06 & 21.01 & 27.35 & 32.54 & 40.20 & 27.43 \\
       & Quest & 20.13 & 24.78 & 31.92 & 43.93 & 59.44 & 36.04 & 16.41 & 20.68 & 27.02 & 37.37 & 51.37 & 30.57 \\
       & ArkVale & 20.72 & 26.29 & 35.12 & 47.87 & 63.30 & 38.66 & 16.58 & 20.94 & 28.14 & 38.21 & 53.43 & 31.46 \\
       & LouisKV & 39.27 & 56.79 & 74.58 & 86.36 & 93.91 & 70.18 & 35.39 & 48.11 & 61.12 & 81.27 & 91.19 & 63.42 \\ 
       & \mymethod{} & \textbf{44.38} & \textbf{68.94} & \textbf{84.53} & \textbf{92.45} & \textbf{94.40} & \textbf{76.94} & \textbf{41.70} & \textbf{65.52} & \textbf{80.54} & \textbf{90.51} & \textbf{93.79} & \textbf{74.41} \\ \midrule
      \multirow{6}{*}{\northm{}} & FullCache & \multicolumn{5}{c}{94.62} & 94.62 & \multicolumn{5}{c}{91.19} & 91.19 \\ \cmidrule(l){2-14}
       & KeyDiff & 9.50 & 14.86 & 31.25 & 51.44 & 60.20 & 33.45 & 8.28 & 8.70 & 27.09 & 45.46 & 56.99 & 29.30 \\
       & Quest & 12.23 & 20.18 & 35.48 & 48.02 & 64.83 & 36.15 & 12.12 & 19.14 & 32.83 & 44.16 & 57.62 & 33.17 \\
       & ArkVale & 12.82 & 20.28 & 34.52 & 52.50 & 68.66 & 37.75 & 12.61 & 18.21 & 30.54 & 46.49 & 61.10 & 33.79 \\
       & LouisKV & 23.36 & 51.06 & 74.75 & 84.93 & 91.62 & 65.14 & 14.53 & 20.26 & 61.74 & 79.18 & 86.72 & 52.49 \\ 
       & \mymethod{} & \textbf{34.85} & \textbf{60.72} & \textbf{76.72} & \textbf{88.19} & \textbf{92.70} & \textbf{70.64} & \textbf{36.43} & \textbf{57.83} & \textbf{72.45} & \textbf{83.66} & \textbf{89.26} & \textbf{67.93} \\ \midrule
      \multirow{6}{*}{\cmdr{}} & FullCache & \multicolumn{5}{c}{89.64} & 89.64 & \multicolumn{5}{c}{85.95} & 85.95 \\ \cmidrule(l){2-14}
       & KeyDiff & 12.15 & 17.88 & 29.10 & 48.90 & 58.79 & 33.36 & 10.04 & 15.82 & 22.58 & 40.64 & 54.72 & 28.76 \\
       & Quest & 12.91 & 19.26 & 35.71 & 47.24 & 57.09 & 34.44 & 9.98 & 14.82 & 24.84 & 37.16 & 49.26 & 27.21 \\
       & ArkVale & 12.12 & 17.55 & 27.46 & 45.09 & 58.01 & 32.05 & 9.88 & 14.90 & 20.29 & 33.43 & 46.95 & 25.09 \\
       & LouisKV & \textbf{36.24} & 54.30 & 72.24 & 83.47 & \textbf{88.18} & 66.88 & 26.59 & 39.64 & 60.75 & 74.00 & 83.81 & 56.96 \\ 
       & \mymethod{} & 35.60 & \textbf{56.99} & \textbf{74.32} & \textbf{85.09} & 88.13 & \textbf{68.03} & \textbf{33.09} & \textbf{53.15} & \textbf{67.66} & \textbf{79.84} & \textbf{84.17} & \textbf{63.58} \\ \bottomrule
    \end{tabular}%
  }
\end{table}
\begin{figure}[t]
    \centering
    \includegraphics[width=\linewidth]{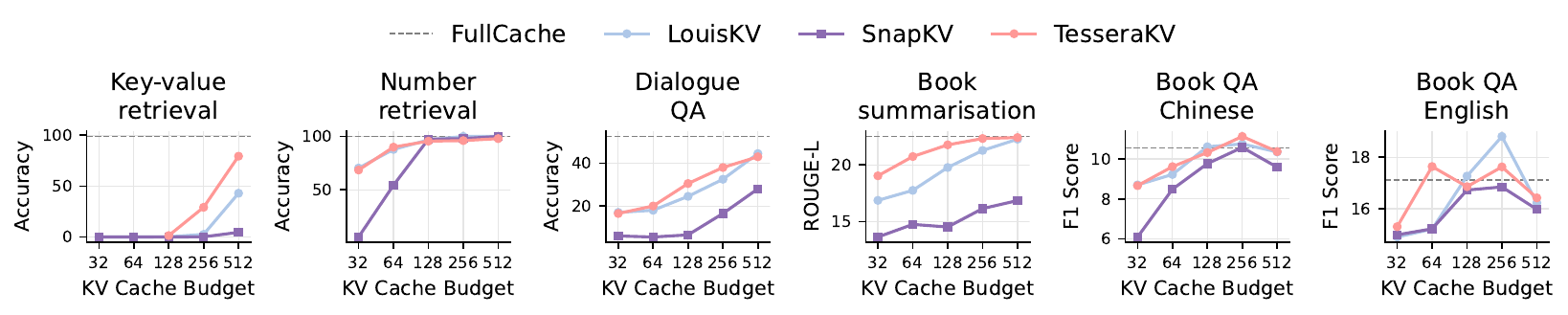}
    \caption{
    Performance comparison of \mymethod{} with LouisKV, SnapKV, and full-cache attention across KV-cache budgets on \textsc{InfiniteBench} using Muse-Glimmer-30B.
    }
    \label{fig:ib-budget}
    \vspace{-2em}
\end{figure}

\paragraph{Models and benchmarks}
We evaluate \museg{}~\citep{museglimmer30b}, \northm{}~\citep{north-mini-code}, and
\cmdr{}~\citep{cmdr-7b}, whose full-attention layers have no positional encoding and whose GQA group widths are $16$, $8$, $4$, respectively. We discuss and evaluate on RoPE models with LLaMA-3.1-8B and Mistral-3-8B in Appendix~\ref{app:limitations}.
Paging applies only to these layers with sliding-window layers remain unchanged. 
For long-context tasks, we use \textsc{ruler}~\citep{ruler} at $8$K and $16$K.
Its $13$ synthetic tasks span needle retrieval, variable tracing, and question answering. 
To complement these synthetic tasks, we evaluate on \textsc{InfiniteBench}~\citep{infinitebench}, whose contexts average over $100$K tokens, and include novels and dialogue scripts. 
For agentic workflows, we use $155$ text-agent tasks from \textsc{Claw-Eval}~\citep{claw-eval}, spanning communication, web search, finance, productivity.

\subsection{Covariance Estimation Accuracy}
\label{sec:exp:estimation-acc}
We ask whether queries available at partition time reliably proxy the decoding
distribution, and whether the answer depends on the task. We consider two
regimes: a balanced subset of \textsc{ruler}, which retrieves from synthetic
contexts, and \textsc{InfiniteBench} book summarisation, which reads natural
prose. For each, we compute the reference covariance $\Sstar$ from decoding
queries and compare it with three pre-decoding $\hSig$ estimators: \textit{Prefill},
\textit{query-mix} with $\mixw$, and \textit{identity ($\mI$) -- no covariance}. See details in Appendix~\ref{app:estimator}.

On \textsc{ruler}, \textit{query-mix} improves both metrics, although they favor
different query-mix weights. Co-retrieval agreement recovers $57.3\%$ of the
distance from $\mI$ ($0.453$) to the reference ($0.579$), peaking at
$\mixw=0.65$; $\mathcal A$ recovers $18.2\%$ of the corresponding distance
from $\mI$ ($0.383$) to the reference ($1$), peaking at $\mixw=0.45$. Neither endpoint helps: the
context-only estimate performs slightly below the identity, while the
question-only estimate falls back toward it. The prefill estimator is likewise
indistinguishable from the identity, indicating that the gain comes from
combining the two query populations. 


\begin{wrapfigure}{r}{0.4\textwidth}
    \vspace{-\intextsep}      
    \vspace{-3em}      
    \centering
    \begin{subfigure}{\linewidth}
        \centering
        \includegraphics[width=\linewidth]{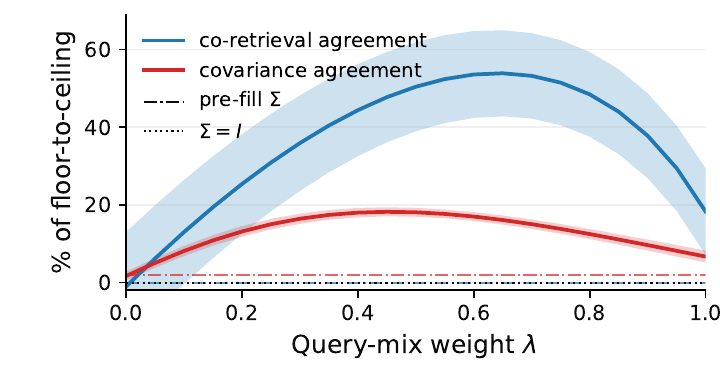}
        \label{fig:mismatch:ruler}
    \end{subfigure} \\
    \vspace{-2.2em}
    \begin{subfigure}{\linewidth}
        \centering
        \includegraphics[width=\linewidth]{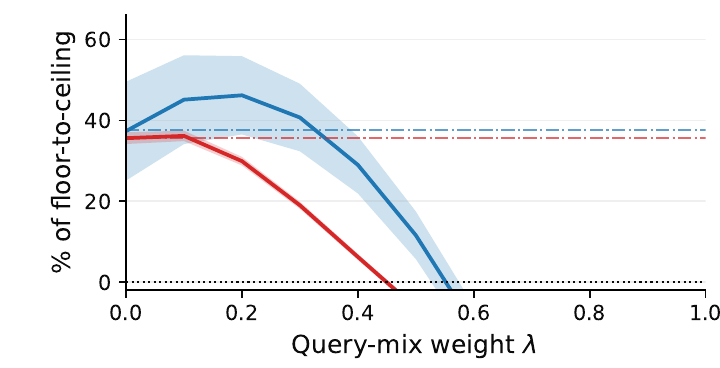}
        \label{fig:mismatch:longbook}
    \end{subfigure}
    \vspace{-3em}
    \caption{Both metrics against the query-mix weight $\mixw$, as a share of the range from
    $\Sq = \mI$ to the reference $\Sstar$
    . Signal carried by flips with the task:
    pre-fill sits on the floor for \textsc{ruler} (top) and up for \textsc{Infinite Bench} (bottom).}
    \vspace{-2em}

    \label{fig:metric-mismatch-both}
\end{wrapfigure}

On \textsc{InfiniteBench}, the pattern reverses. The context-only estimate
already recovers $37\%$ of the distance to the reference. Adding question
queries yields a further improvement, reaching $50\%$ at $\mixw=0.2$, but
becomes harmful thereafter; at $\mixw=0.6$, performance falls below the
identity.

This contrast reflects the different roles of the context and question in the two tasks.
In \textsc{ruler}, the question determines what needs to be retrieved, so
question queries carry the relevant signal while the synthetic context
contributes little. In book summarisation, the decoding queries are directed
at the document itself, so the context carries much more of the relevant
signal. The two metrics also need not move together: an estimator can be a
poor approximation of the full covariance while still capturing the geometry
relevant to partitioning if its error lies primarily in directions along which
the keys vary little.
The same analysis across additional tasks is reported
in Appendix~\ref{app:task-sweeps}.

\subsection{Main Result}
\label{sec:main}

\begin{figure}[t]
    \centering
    \includegraphics[width=\linewidth]{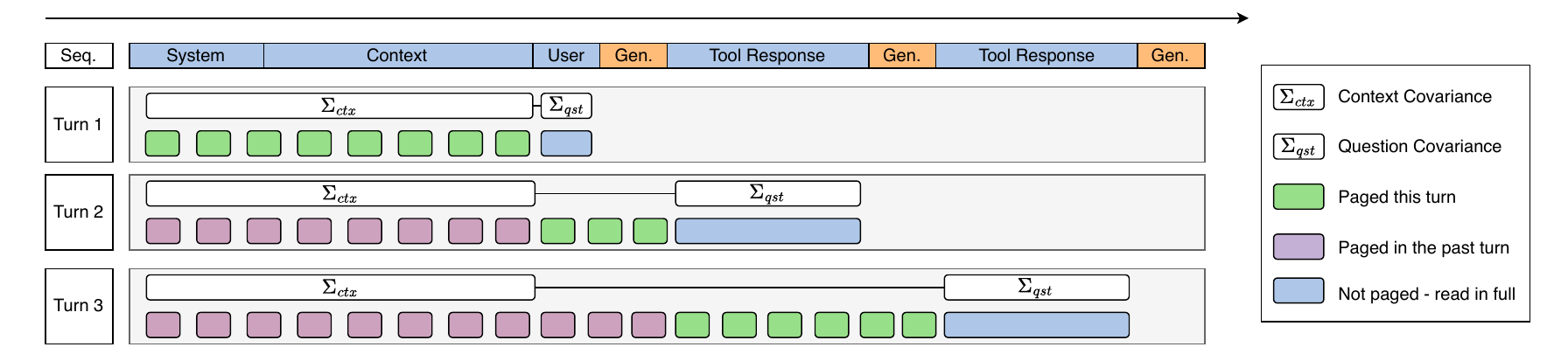}
\caption{
    \mymethod{} across the turns of an agentic workload. Each turn partitions only
    the tokens appended since the last turn and keeps the pages earlier turns built;
    its own question stays unpaged and is read in full. $\Sigma_{ctx}$ is estimated
    once from the opening context, $\Sigma_{qst}$ once per turn.
    }
    \label{fig:07:flow}
\end{figure}
\begin{table}[!t]
\caption{
Evaluation on \textsc{Claw-Eval} using a $131{,}072$-token context window across $155$ agentic tasks, with three trials per task on \museg{}. A task is considered successful only if all three trials exceed the rubric threshold. Results report task success rates by difficulty. KV budget $\budget$ denotes the number of cached tokens retrieved per KV head at each decoding step.
\emph{Overall} is task-weighted performance
(Easy: $69$, Medium: $46$, Hard: $40$). 
Bold indicates the best result in each column.}

  \label{tab:claweval}
  \centering

\setlength{\tabcolsep}{3.5pt}

\renewcommand{\arraystretch}{1.1}
  \resizebox{\textwidth}{!}{%
    \begin{tabular}{@{}lcccc|cccc|cccc|cccc@{}}
      \toprule
       & \multicolumn{4}{c}{Easy} & \multicolumn{4}{c}{Medium}
                              & \multicolumn{4}{c}{Hard} & \multicolumn{4}{c}{Overall} \\
      \cmidrule(lr){2-5} \cmidrule(lr){6-9} \cmidrule(lr){10-13} \cmidrule(l){14-17}
      KV budget $\budget$ & $64$ & $128$ & $256$ & $512$ & $64$ & $128$ & $256$ & $512$ & $64$ & $128$ & $256$ & $512$ & $64$ & $128$ & $256$ & $512$ \\ \midrule
      FullCache & \multicolumn{4}{c}{73.91} & \multicolumn{4}{c}{86.96} & \multicolumn{4}{c}{70.00} & \multicolumn{4}{c}{76.77} \\ \cmidrule(l){1-17}
      Arkvale & 24.64 & 42.03 & 66.67 & 72.46 & 2.17 & 34.78 & 69.57 & 80.43 & 5.00 & 17.50 & 57.50 & 60.00 & 12.90 & 33.55 & 65.16 & 71.61 \\
      LouisKV & 28.99 & 47.83 & \textbf{73.91} & 69.57 & 17.39 & 45.65 & 73.91 & \textbf{84.78} & 10.00 & 30.00 & 45.00 & 62.50 & 20.65 & 42.58 & 66.45 & 72.26 \\
      $\text{\mymethod{}}$ & \textbf{59.42} & \textbf{71.01} & \textbf{73.91} & \textbf{75.36} & \textbf{63.04} & \textbf{82.61} & \textbf{80.43} & 82.61 & \textbf{42.50} & \textbf{55.00} & \textbf{62.50} & \textbf{65.00} & \textbf{56.13} & \textbf{70.32} & \textbf{72.90} & \textbf{74.84} \\
      
      \bottomrule
    \end{tabular}%
  }
\end{table}

This section presents the main evaluation results of \mymethod{} across
long-context retrieval and understanding tasks, followed by agentic
workloads. We conclude with profiling results that quantify the hardware
efficiency gains and the overhead of estimating the query covariance.

\paragraph{Long-Context Tasks}
\label{par:downstream}

In~\Cref{tab:main16k}, we compare \mymethod{} against eviction and retrieval
baselines on \textsc{ruler}-8K and 16K. Retrieval methods including Quest,
Arkvale, and LouisKV, generally outperform eviction methods like KeyDiff,
showing that permanently discarding KV entries can degrade accuracy.

Notably, LouisKV groups keys by cosine similarity, corresponding to the
identity (no-covariance) geometry considered in our analysis. \mymethod{}
instead uses the query covariance to adapt the key geometry to the
co-retrieval structure of the head. It outperforms all baselines in most
settings. On \textsc{ruler}-16K with Muse-Glimmer-30B, \mymethod{} achieves
an average accuracy of $74.41$, improving over LouisKV's $63.42$ by
$10.99$ points, averaged across KV budgets.
We present the \textsc{InfiniteBench} results in~\Cref{fig:ib-budget}.
\mymethod{} matches or outperforms LouisKV in six tasks at almost all budgets,
with the largest gains on book summarisation. This result is consistent with
our covariance analysis: for natural-language documents, the context itself
provides a strong signal for the decoding-query geometry, making it useful
for constructing the partition.

\paragraph{Agentic Workloads}
\label{par:agentic}

We evaluate on \textsc{Claw-Eval} in~\Cref{tab:claweval}, extending
\mymethod{} to multi-turn tasks by estimating $\Sigma_{ctx}$ once from the
opening context and re-estimating only $\Sigma_q$ each turn. We partition only
newly appended tokens while retaining pages from earlier turns, avoiding
re-partitioning the full context as shown in \cref{fig:07:flow}. 
\mymethod{} substantially outperforms ArkVale and LouisKV at tight KV-cache budgets, achieving an overall success rate of $70.32$ versus $33.55$ and $42.58$ at $\budget=128$, respectively. From $128$ to $512$, its performance remains relatively stable, with only a small gap to full cache, while both ArkVale and LouisKV degrade more substantially. At $512$, the methods narrow the gap, although Hard tasks remain below full-cache accuracy. Comparisons with full repartitioning and AgentKV are reported in Appendix~\ref{app:claweval-budget}.

We report the end-to-end cost of completing \textsc{Claw-Eval} tasks, including index construction and decoding with the KV cache offloaded to CPU memory, in Appendix~\ref{app:offload}.

\paragraph{Inference Throughput}

\begin{figure}[t]
    \centering
    \includegraphics[width=0.9\linewidth]{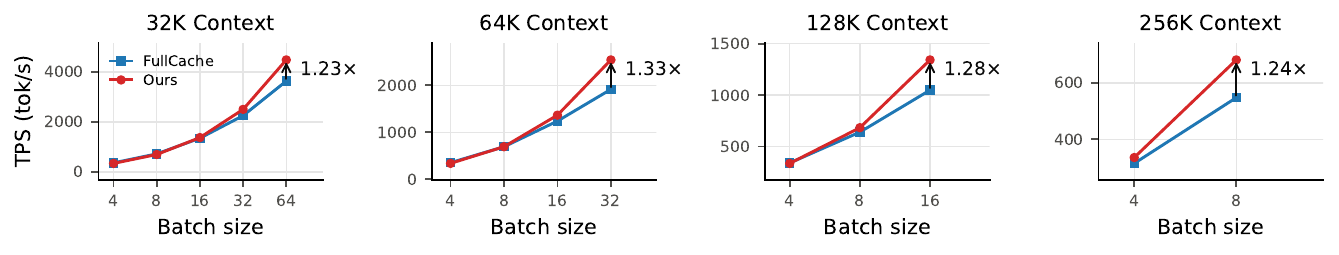}
    \vspace{-1em}
    \caption{Decoding throughput against batch size at four context lengths up to device memory limit.}
    \label{fig:throughput-batch}
    \centering
    \includegraphics[width=\linewidth]{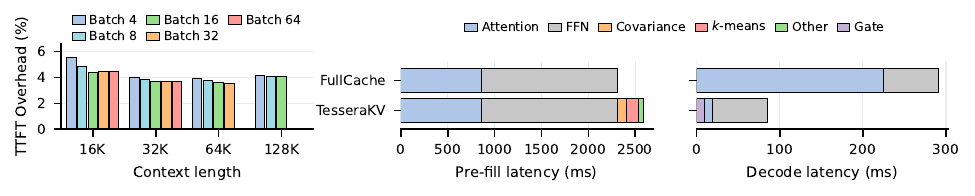}
    \vspace{-2em}
    \caption{Time-to-first-token overhead and profiled latency components of \mymethod{} relative to full-cache attention. \emph{Left:}  TTFT overhead across context lengths and batch sizes. \emph{Middle and right:} prefill and decode component times of one global attention layer at $64$k context and batch $32$. Decode times are summed over $512$ steps. Component times are summed without asynchronous overlap.}
    \label{fig:latency}
    \vspace{-1em}
\end{figure}

\Cref{fig:throughput-batch} reports decoding throughput across batch sizes and context lengths. Each request consists of the context tokens followed by a $256$-token question, and generates $512$ tokens. At the largest batch size that both configurations can support on the B200 with SGLang and FlashAttention 4, \mymethod{} achieves $1.23$--$1.33\times$ the throughput of full-cache attention across context lengths from $32$k to $256$k. At batch size $8$, \mymethod{} retains $97.4\%$ of its throughput as the context grows from $8$k to $256$k, while full-cache throughput drops by $26.5\%$.
Our implementation reduces paging overhead through batched matrix multiplication for page scoring, fused Triton kernels for page selection and address packing, and sparse attention. Implementation details are provided in Appendix~\ref{app:inference-impl}.\

The left panel of \cref{fig:latency} reports TTFT. Relative to full-cache attention, \mymethod{} adds around $3.5$--$4.0\%$ from $32$k to $128$k, with little variation across batch sizes. The remaining panels profile one layer with global attention at $64$k context and batch size $32$. When component times are summed without asynchronous overlap, page selection and sparse attention together take one-twelfth the time of full attention during decoding.
During prefill, covariance estimation accounts for $4\%$ of the summed component time. Additional breakdowns appear in Appendix~\ref{app:latency}.

\subsection{Ablation study}
\label{sec:exp:ablation}
To isolate the contributions of the covariance-based partition and gate, we cross three partition choices (positional, $\Sq=\mI$, and the \textit{query-mix}) with three gates (none, the spread measured under $\mI$, and under $\Sq$) in Appendix~\ref{app:sigma-gate}. The covariance-based setting generally performs best in both components and jointly. We further ablate sharing a partition across GQA groups, the form and weight of the spread correction, and the query-mix weight across workloads. Details are provided and Appendix~\ref{app:ablation}.

\section{Conclusion}
\label{sec:conclusion}

We propose \mymethod{}, a KV paging, it formulates partitioning and page scoring around the co-retrieval structure of attention. \mymethod{} reduces KV bandwidth requirements during decoding for long-context and agentic workloads, while achieving state-of-the-art performance.
  

\newpage
\section*{AI Disclosure}
In this work, we used generative AI tools to assist with polishing the writing and editing code development, and launching experiments at large scale. We did not use generative AI tools to generate research ideas or synthetic datasets, or to formulate mathematical claims. We take full responsibility for the final content of this work, including all text, claims, and artifacts produced with the assistance of generative AI.

\bibliography{iclr2027_conference}
\bibliographystyle{iclr2027_conference}

\newpage
\appendix
\section{Notation}
\label{app:notation}

\Cref{tab:app:notation:matrix} collects the vectors, matrices and sets used in this paper,
and \Cref{tab:app:notation:dim} the dimensions and scalars.
\begin{table}[ht]
    \caption{Notation of vectors, matrices, mappings, and sets in this paper.}
    \vspace{0.5em}
    \label{tab:app:notation:matrix}
    \centering
    \begin{small}
    \begin{tabular}{ll}
        \toprule
        Notation        & Description                                                                  \\ \midrule
        $\query$        & A decoding query of a head; $\query_t$ is the query of step $t$               \\
        $\key{i}$       & Key of cached position $i$                                                    \\
        $\cent{p}$      & Centroid of page $p$, the mean of its keys in the raw key space               \\
        $\keyw{i}$      & Transformed key, $\keyw{i} = \Sqhalf \key{i}$                                 \\
        $\centw{p}$     & Transformed centroid, also the mean of $\{\keyw{i}\}_{i \in \clus{p}}$        \\
        \midrule
        $\Sq$           & Query covariance of a head, \cref{sec:key-retrieval}                         \\
        $\Sqhalf$       & Unique symmetric positive semidefinite square root of $\Sq$                   \\
        $\phimap$       & Key transform $\phimap(\key{}) = \Sqhalf \key{}$; uses $\Sbar^{1/2}$ for GQA \\
        $\Sctx$         & Covariance of the queries produced by the context block                       \\
        $\Sqst$         & Covariance of the queries produced by the question block                      \\
        $\hSig_\mixw$   & Query-mix estimator of $\Sq$, \cref{eq:question-mix}                        \\
        $\Sstar$        & Reference covariance estimated from the decoding queries                         \\
        $\Sbar$         & Covariance pooled over a KV group, one per KV head, \cref{eq:avg-graph}       \\
        $\mI$           & Identity matrix; $\Sq = \mI$ recovers plain cosine similarity                 \\
        \midrule
        $\pmap$         & Partition map, from a key index to a page index                               \\
        $\mathcal P$    & The partition $\{\clus{1}, \ldots, \clus{\npage}\}$                            \\
        $\clus{p}$      & Page $p$, the set of key indices it holds                                     \\
        $\Rset$         & Resident keys, the first $\nsink$ positions                                   \\
        $\Cset$         & Clustered keys, the positions the partition is built over                     \\
        $\Oset$         & Open tail: the question and what has been generated, always read              \\
        $\Sel$          & Key slots admitted by the budget walk at the current step                     \\
        $\wset$         & Working set actually attended over, $\Rset \cup \Sel \cup \Oset$              \\ \bottomrule
    \end{tabular}
    \end{small}
\end{table}

\begin{table}[ht]
    \caption{Notation of dimensions and scalars in this paper.}
    \vspace{0.5em}
    \label{tab:app:notation:dim}
    \centering
    \begin{small}
    \begin{tabular}{ll}
        \toprule
        Notation      & Description                                                              \\ \midrule
        $\dhead$      & Head dimension, shared by the query and the key                          \\
        $\nkeys$      & Context keys per layer and KV head                                      \\
        $\gqa$        & GQA group width: query heads per KV head, indexed by $g$                 \\
        $\npage$      & Number of pages the clustered keys are partitioned into                  \\
        $\csize$      & Target page size, $\npage = \max(\lceil |\Cset| / \csize \rceil,1)$     \\
        $\budget$     & Retrieval budget for clustered keys per KV head per decoding step       \\
        $\nsink$      & Length of the resident prefix held as attention sinks                    \\
        $\mixw$       & Query-mix weight of \cref{eq:question-mix}                               \\
        $\kgate'$     & Coefficient of the linear spread correction $\kgate'\spread{p}$        \\
        \midrule
        $\logit{ti}$  & Attention logit of key $i$ under the query of step $t$                   \\
        $a_{ti}$      & Normalised attention score of key $i$ at step $t$                        \\
        $g_{tp}$      & Centroid logit of page $p$ under query $t$, \cref{eq:decomp}              \\
        $\delta_{ti}$ & Deviation of member $i$ from its page's centroid score, \cref{eq:decomp} \\
        $U_t(p)$     & Average unnormalized attention mass of page $p$ at step $t$              \\
        $\spread{p}^2$& Mean within-page logit residual variance, \cref{eq:spread}              \\
        $\spread{p}$  & Square root of the mean residual variance used in the linear gate        \\
        $\gscore{p}$ & Page score aggregated across the query heads of a KV group              \\
        $\rho_{ij}$   & Co-retrieval correlation of keys $i$ and $j$, \cref{eq:corr}             \\
        $\mathcal A$  & Covariance agreement of an estimator, \cref{eq:cov-agreement}            \\
        $J$           & Partition objective, \cref{eq:objective}                                 \\ \bottomrule
    \end{tabular}
    \end{small}
\end{table}

\section{Algorithm}
\label{app:algorithm}

\Cref{alg:prefill} details the construction of partitions and \Cref{alg:decode} details the page retrieval process. The partition is
built once per prefill, for each paged layer and KV head, at the end of the question block; 
the gate is evaluated at each decoding step by query heads.
Empty pages after the final Lloyd assignment are skipped when computing centroids and spreads and masked during decoding.


\begin{algorithm}[h]
\caption{\mymethod{} pre-fill: constructing the partition.}
\label{alg:prefill}
\begin{algorithmic}[1]
\Require Context keys $\mK \in \mathbb{R}^{\nkeys \times \dhead}$; context queries $\{\mathbf{Q}^{(g)}_\mathrm{ctx}\}_{g=1}^{\gqa}$; question queries $\{\mathbf{Q}^{(g)}_\mathrm{qst}\}_{g=1}^{\gqa}$;
         query-mix weight $\mixw$; target page size $\csize$; sinks $\nsink$
\Ensure  Partition $\pmap$, centroids $\{\cent{p}\}$, dispersions $\{\spread{p}\}$,
         sizes $\{|\clus{p}|\}$, slot table
\Statex
\For{$g = 1$ \textbf{to} $\gqa$}
    \Comment{\emph{per-query-head} query statistics}
    \State $\Sctx^{(g)} \gets \operatorname{Cov}(\mathbf{Q}^{(g)}_\mathrm{ctx})$
    \State $\Sqst^{(g)} \gets \operatorname{Cov}(\mathbf{Q}^{(g)}_\mathrm{qst})
             \cdot \operatorname{tr}\Sctx^{(g)} / \operatorname{tr}\Sqst^{(g)}$
           \Comment{trace-match to $\Sctx^{(g)}$}
    \State $\Sq^{(g)} \gets (1-\mixw)\,\Sctx^{(g)} + \mixw\,\Sqst^{(g)}$
\EndFor
\State $\Sbar \gets \frac{1}{\gqa} \sum_{g=1}^{\gqa} \Sq^{(g)}$
       \Comment{the shared metric of \Cref{sec:gqa}: average}
\State $\phimap(\vk) \gets \Sbar^{1/2}\vk$
\Statex
\State $\Rset \gets \{1,\dots,\nsink\}$;\quad $\Cset \gets \{\nsink+1,\dots,\nkeys\}$
       \Comment{sinks are excluded from partition}
\State $\npage \gets \max\!\big(\lceil |\Cset| / \csize \rceil,\, 1\big)$
\State $\pmap \gets \textsc{Lloyd}\big(\{\phimap(\key{i})\}_{i \in \Cset},\, \npage\big)$
       \Comment{$10$ iterations, Euclidean k-mean}
\Statex
\For{$p = 1$ \textbf{to} $\npage$}
    \State $\cent{p} \gets \frac{1}{|\clus{p}|}\sum_{i \in \clus{p}} \key{i}$
    \State $\spread{p}^2 \gets \frac{1}{|\clus{p}|}\sum_{i \in \clus{p}}
           \frac{(\key{i}-\cent{p})^{\!\top}\Sbar\,(\key{i}-\cent{p})}{\dhead}$
    \State store $\clus{p}$'s indices in slots by \emph{decreasing} position
\EndFor
\State \Return $\pmap$, $\{\cent{p}\}$, $\{\spread{p}\}$, $\{|\clus{p}|\}$, slot table
       \Comment{$\Sbar$ and $\phimap$ are discarded}
\end{algorithmic}
\end{algorithm}

\begin{algorithm}[t]
\caption{\mymethod{} decoding step: selecting the working set.}
\label{alg:decode}
\begin{algorithmic}[1]
\Require Queries $\{\query_g\}_{g=1}^{\gqa}$;
         centroids $\{\cent{p}\}$, dispersions $\{\spread{p}\}$, sizes $\{|\clus{p}|\}$, slot table;
         budget $\budget$; coefficient $\kgate'$; sink set $\Rset$; open tail $\Oset$
\Ensure  Working set $\wset$
\Statex
\For{$g = 1$ \textbf{to} $\gqa$, \; $p = 1$ \textbf{to} $\npage$}
    \State $\gscore{g,p} \gets
           \dfrac{\query_g \cdot \cent{p}}{\sqrt{\dhead}} \;+\; \kgate'\spread{p}$
\EndFor
\State $\gscore{g,p} \gets -\infty$ \textbf{ for all } $p$ \textbf{ with } $|\clus{p}| = 0$
\Statex
\State $\gscore{p} \gets \frac{1}{\gqa}\sum_{g=1}^{\gqa}
       \big[\operatorname{softmax}_{p'}(\gscore{g,p'})\big]_p$
       \Comment{group-consistent reduce: a page must be wanted by \emph{many} heads}
\Statex
\State $(o_1,\dots,o_\npage) \gets$ the page indices ordered by decreasing $\gscore{p}$
       \Comment{$\clus{o_j}$ is the $j$-th ranked page}
\State $\textit{spent} \gets 0$;\quad $\Sel \gets \emptyset$
\For{$j = 1$ \textbf{to} $\npage$}
    \Comment{budget walk in \emph{keys}}
    \State $t \gets \min\!\big(|\clus{o_j}|,\; \max(\budget - \textit{spent},\, 0)\big)$
    \State $\Sel \gets \Sel \cup \{\text{first } t \text{ slots of } \clus{o_j}\}$;\quad
           $\textit{spent} \gets \textit{spent} + t$
\EndFor
\Statex
\State $\wset \gets \Rset \;\cup\; \Sel \;\cup\; \Oset$
       \Comment{$\Oset$ is the question plus what has been generated: read in full}
\State \Return $\wset$
       \Comment{written to the request's index table}
\end{algorithmic}
\end{algorithm}

\section{Co-retrieval on Natural Text}
\label{app:coretrieval}

\Cref{fig:03:prefill} compares two ways of predicting whether a pair of keys
has correlated attention across decoding queries: the plain cosine used by
prior work, and the $\Sq$-weighted cosine of \Cref{eq:corr}. This section
describes how each quantity in the figure is computed.

\paragraph{Data.}
We use six SlimPajama domains~\citep{slimpajama} --- ArXiv, Books, C4,
CommonCrawl, GitHub, and Wikipedia --- with $41$ to $45$ documents per model,
each truncated to its first $16$k tokens. Measurements are taken on the global
full-attention layers only.

\paragraph{Query populations.}
A document is passed through the model once. The queries emitted while reading
the $16$k-token context give the estimate $\hSig$, the covariance of those
context queries. The model then decodes $256$ further tokens, and the queries
at those positions are held out: they never enter $\hSig$ and are the only
queries used to evaluate the predictions.

\paragraph{The measured quantity.}
For each (document, layer), we draw $8{,}192$ key pairs $(i,j)$ uniformly from
the cached keys, with $i\neq j$. For each pair, the \emph{empirical
co-retrieval} is the correlation of their attention logits across the held-out
queries,
$\rho^{\mathrm{emp}}_{ij} =
\operatorname{Corr}_t(\logit{ti},\logit{tj})$,
computed separately for each query head. We compare this quantity with two
predictors. The plain cosine is
$\key{i}^\top\key{j}/(\lVert\key{i}\rVert\lVert\key{j}\rVert)$, which
\Cref{eq:corr} reduces to when $\Sq=\mI$. The $\Sq$-weighted cosine is
\Cref{eq:corr} evaluated at $\hSig$.

\paragraph{The score.}
Within each (document, layer, head), we compute the Pearson correlation between
a predictor and $\rho^{\mathrm{emp}}$ across the $8{,}192$ pairs, yielding one
agreement $\bar r$ per head per document. These values are averaged over
documents before forming the histogram, so each unit in \Cref{fig:03:prefill}
is a (layer, head) pair: $416$ units for \museg{}, $384$ for \northm{}, and
$256$ for \cmdr{}. 


\subsection{Predicting Attention-Logit Covariance}
\label{app:coretrieval-cov}

We repeat the analysis of \cref{fig:03:prefill} for attention-logit covariance, which retains the magnitude of joint logit fluctuations. For each key pair, \cref{eq:logit-cov} predicts covariance as $\key{i}^{\top}\hSig\key{j}/\dhead$; the identity baseline predicts $\key{i}^{\top}\key{j}/\dhead$. We compare both with the empirical covariance over the same held-out decoding queries used for the correlation analysis.
For each document, layer, and head, we compute the Pearson correlation between predicted and empirical covariances across sampled key pairs, then average this value over documents. This measures how well each predictor distinguishes key pairs with different covariances, rather than its calibration in absolute units.

\begin{figure}[h]
    \centering
    \includegraphics[width=\linewidth]{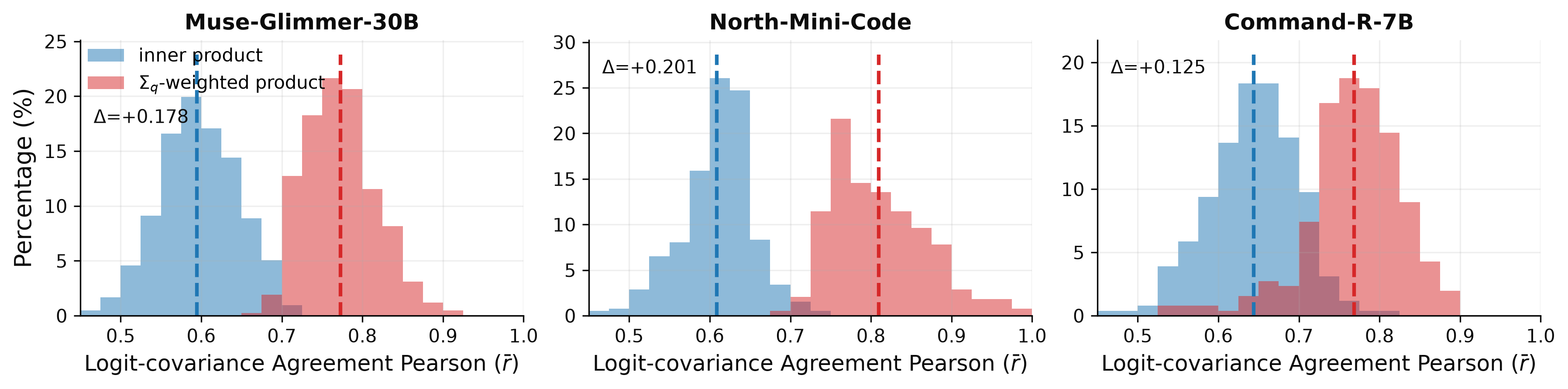}
    \caption{Agreement between predicted and empirical attention-logit covariances on SlimPajama. Each panel shows the distribution over (layer, head) pairs of Pearson correlations $\bar r$ across sampled key pairs, averaged over documents from six domains. The predictors use either the prefill query covariance in~\eqref{eq:logit-cov} or the identity matrix, which gives the scaled raw inner product. Larger $\bar r$ indicates better agreement; dashed lines mark the mean for each predictor.}
    \label{fig:03:prefill-cov}
\end{figure}

\Cref{fig:03:prefill-cov} shows higher agreement with the estimated query covariance than with the identity baseline in all three models. The mean gains are $0.178$, $0.201$, and $0.125$ for \museg{}, \northm{}, and \cmdr{}, respectively. The corresponding mean gains for logit correlation are $0.134$, $0.179$, and $0.125$. Thus, the benefit of query covariance persists when the magnitude of logit co-fluctuation is retained.

\section{Detailed Experimental Setup}
\label{app:exp}

\paragraph{Configuration.}
We estimate query covariance from context and question queries using \cref{eq:question-mix}, retain the first four positions as attention sinks~\citep{xiao2023streamingllm, what-are-you-sinking, secondary-attention-sinks}, and share one partition per KV group under $\Sbar$. All experiments use 10 $k$-means iterations. \Cref{tab:configs} lists the query-mix weight and gate coefficient for each benchmark; their effects are examined in \cref{sec:exp:estimation-acc}, Appendix~\ref{app:gate-form} and Appendix~\ref{app:task-sweeps}. \textsc{Claw-Eval} uses the same configuration as \textsc{InfiniteBench}.

\paragraph{KV-cache budge.}
The budget $\budget$ counts only the context keys selected per KV head at each decoding step. For all methods, the four attention-sink tokens, the question block, and all generated tokens are always attended in full and excluded from the budget. Thus, all methods attend to the same uncompressed tokens and differ only in their selection of $\budget$ context keys, ensuring matched per-step KV traffic across \textsc{ruler}, \textsc{InfiniteBench}, and \textsc{Claw-Eval}.

For the multi-turn \textsc{Claw-Eval} comparison, we extend LouisKV by treating the accumulated context at each turn as a new prefill and rebuilding its partition. This follows the same repartitioning schedule as the $\text{\mymethod{}}_{\mathrm{Repartition}}$ variant.


\paragraph{Hyperparameter selection.} We calibrate both hyperparameters offline on \museg{}, using two small subsets: a balanced \textsc{Ruler}-16K subset with $10$ sequences per task for synthetic retrieval, and $10$ \textsc{InfiniteBench} Book Summarization documents for natural prose. Pooling the decoding queries across these 10 sequences provides sufficient data to estimate a stable reference covariance for each head (Appendix~\ref{app:estimator}). We select the query-mix weight $\mixw$ using the two intrinsic metrics introduced in \Cref{sec:cov-estimation}. We select the gate coefficient $\kappa'$ based on end-task scores from a coefficient sweep on the same two workloads (Appendix~\ref{app:gate-form}). Both hyperparameters exhibit a systematic difference between synthetic retrieval and natural prose, rather than requiring task-specific optima. We therefore use a single setting for each regime: $\mixw=0.5$ and $\kappa'=6$ for all \textsc{Ruler} tasks, and $\mixw=0.2$ and $\kappa'=4$ for all \textsc{InfiniteBench} tasks in \Cref{tab:configs}. We keep these settings fixed, without further tuning, for \northm{} and \cmdr{} at every budget. We also apply the natural-text setting to \textsc{Claw-Eval}.

\begin{table}[h]
  \caption{Configuration per benchmark. $\mixw$ is the query-mix weight of \Cref{eq:question-mix} and
  $\kappa'$ the coefficient of the linear spread bonus variant of \Cref{eq:spread-gate}.}
  \vspace{0.5em}
  \label{tab:configs}
  \centering
  \small
  \begin{tabular}{@{}llccll@{}}
    \toprule
    Benchmark & Arm & $\mixw$ & $\kappa'$ & Gate & $k$-means \\ \midrule
    \textsc{ruler} $8$K / $16$K & \mymethod{}             & $0.5$ & $6$ & linear & euclidean \\
    \textsc{InfiniteBench}      & \mymethod{}             & $0.2$ & $4$ & linear & euclidean \\ 
    \multirow{1}{*}{\textsc{Claw-Eval}}
                                & \mymethod{}             & $0.2$ & $4$ & linear & euclidean \\
    \bottomrule
  \end{tabular}
\end{table}

\paragraph{\textsc{Claw-Eval} task specification.}
The benchmark ships $300$ tasks: $161$ text agent tasks (\texttt{T*}), $38$ that drive a
simulated user (\texttt{C*}), and $101$ multimodal ones (\texttt{M*}). We evaluate the
\texttt{T*} family less the six tasks carrying an attachment (\texttt{T056}, \texttt{T057},
\texttt{T058}, \texttt{T072} attach an image; \texttt{T091} and \texttt{T096} a text or CSV
file), leaving $155$.

\subsection{Estimator quality}
\label{app:estimator}
The sweeps in \Cref{fig:metric-mismatch-both} evaluate \museg{} on the $13$ \emph{NoPE} layers. The unit of measurement is a \emph{(sequence, layer) cell}; \Cref{alg:estimator-quality} defines how the two metrics are computed within each cell. 

\begin{algorithm}[h]
\caption{One (sequence, layer) cell. $T$ is the number of decode steps that sequence produced,
$P$ is the number of sampled key pairs, and $a$ ranges over the query heads.}
\label{alg:estimator-quality}
\begin{algorithmic}[1]
  \State $\hSig_\mixw \gets \big((1-\mixw)\,\Sctx + \mixw\,\Sqst\big)$, trace-matched
         \Comment{\Cref{eq:question-mix}, from this sequence's own pre-fill}
  \State draw $P$ pairs $(i, j)$, $i \neq j$, uniformly over the cached keys
         \Comment{once per cell}
  \For{each pair $(i, j)$ and query head $a$}
    \State $\rho^{\mathrm{emp}}_{ij} \gets \operatorname{Corr}_{t \le T}\big(\logit{ti},\, \logit{tj}\big)$
           \Comment{over the decode queries}
    \State $\hat\rho_{ij} \gets \key{i}^\top \hSig_\mixw \key{j} \,/\, \sqrt{\key{i}^\top \hSig_\mixw \key{i} \cdot \key{j}^\top \hSig_\mixw \key{j}}$
           \Comment{\Cref{eq:corr}}
  \EndFor
  \State co-retrieval agreement $\gets \operatorname{mean}_a \operatorname{Corr}_{(i,j)}\big(\hat\rho,\, \rho^{\mathrm{emp}}\big)$
         \Comment{over the $P$ pairs}
  \State $\mathcal{A} \gets \operatorname{mean}_a \big\langle \hSig_\mixw, \Sstar \big\rangle_F \big/ \big(\lVert\hSig_\mixw\rVert_F \lVert\Sstar\rVert_F\big)$
\end{algorithmic}
\end{algorithm}

For each cell, we sample $P=8{,}192$ KV key pairs once and reuse them across all decode steps. The $T$ decode steps provide the samples for estimating empirical correlations. We use $128$ decode tokens per document on \textsc{InfiniteBench} and $5$--$15$ on \textsc{ruler}, depending on answer length, following KVPress~\citep{ea}. Both metrics are computed in a single pass over the same captures.
We sample ten sequences from each of \textsc{ruler}'s $13$ tasks and ten documents from book summarisation.
For each sequence, the $13$ layers of the model share the same document and captured queries. We first average across layers to obtain one value per sequence, then report $\pm1$ standard error across sequences.

For each layer and query head, we estimate the reference covariance $\Sstar$ from decoding queries, centred by their pooled mean, across the rows of the task. On \textsc{ruler}, this pools all $130$ rows, yielding approximately $1{,}328$ queries per head.
On book summarisation, it pools the $10$ documents, yielding $7{,}361$ queries per head.

Because $\Sstar$ is itself a finite-sample estimate, the two metrics have different interpretations. For \emph{co-retrieval agreement}, the target is each row's own empirical correlation rather than $\Sstar$; a covariance estimated from other rows can therefore outperform $\Sstar$ and exceed $100\%$. In contrast, \emph{covariance agreement} $\mathcal{A}$ measures agreement directly with $\Sstar$ and is bounded by $1$.

\begin{figure}[t]
    \centering
    \includegraphics[width=\linewidth]{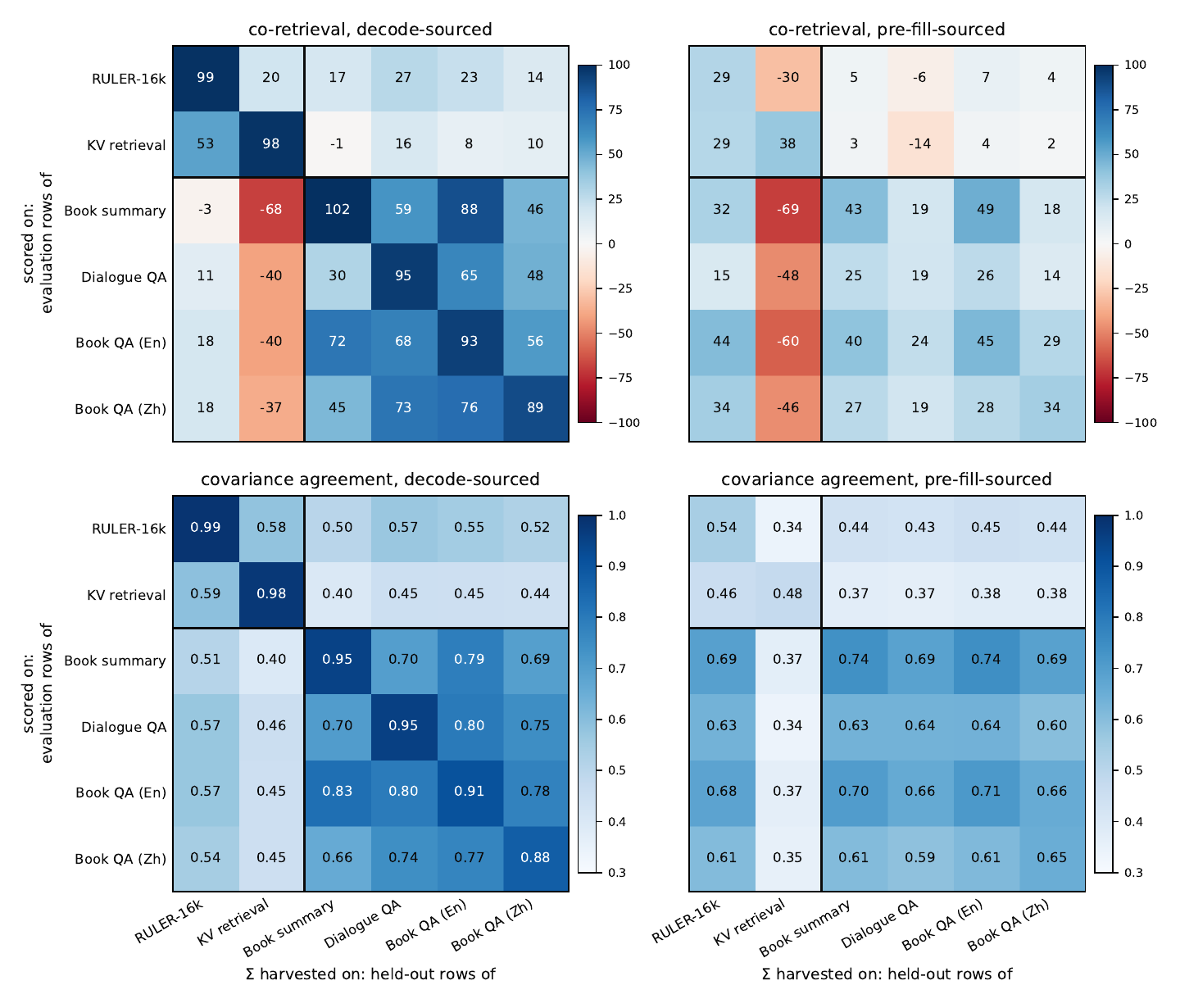}
\caption{Transfer between tasks. Each cell estimates an offline $\Sq$ from the column task and evaluates it on the row task. \textbf{(Top)} Co-retrieval agreement, normalized to the evaluated task's range from $\Sq=\mI$ to its reference; $0$ indicates no improvement over Identity, while values may exceed $100\%$. \textbf{(Bottom)} Covariance agreement. The left and right columns use decoding and pre-fill queries, respectively. The diagonal denotes the held-out within-task setting, where $\Sq$ is estimated from an independent sample of the same task.}

    \label{fig:transfer-grid}
\end{figure}

\paragraph{Sequence Geometry.}
Using $n$ sequences raises the question of whether the results depend on the sampled sequences. We therefore conduct the transfer experiment in \Cref{fig:transfer-grid}: each cell estimates $\Sq$ from one task's rows and evaluates it on another's. The \emph{diagonal} thus measures within-task transfer, where the covariance is estimated from one set of rows and evaluated on disjoint rows from the same task.

The results depend strongly on the query source. When estimated from decoding queries, the diagonal recovers $89\%$--$102\%$ of the identity-to-reference range for co-retrieval agreement and $88\%$--$99\%$ for covariance agreement. In contrast, covariances estimated from pre-fill queries recover only $19\%$--$45\%$. Thus, pre-fill geometry varies substantially across rows within a task, whereas decoding geometry is relatively stable. A covariance estimated from disjoint rows therefore remains close to one estimated on the evaluated rows, suggesting that decoding-query geometry is primarily a property of the task rather than the particular sample.

The off-diagonal entries measure cross-task transfer and show a different pattern. Among the four natural-text tasks, covariances transferred from another task recover $30\%$--$88\%$ of the identity-to-reference range. In contrast, using \textsc{KV Retrieval} as the source is consistently harmful, yielding $-68\%$ to $-37\%$, below the identity baseline. Thus, synthetic-filler retrieval and natural-prose reading induce distinct decoding-query subspaces, such that a covariance learned from one can degrade performance on the other.

\section{Ablation Study}
\label{app:ablation}
\subsection{The GQA group}
\label{app:gqa}
\begin{table}[h]
\caption{
Per-query-head versus shared partitioning on a balanced subset of \textsc{ruler}-16k, averaged over
five KV budgets and $130$ samples per cell. 
}
  \vspace{0.5em}
  \label{tab:gqa}
  \centering
  \small
  \begin{tabular}{@{}lcccc@{}}
    \toprule
    Model & $\gqa$ & Per query head & Shared, $\Sbar$ & Margin \\ \midrule
    \museg{}                     & 16 & 79.28 & 75.14 & $+4.14$ \\
    \northm{} &  8 & 72.46 & 69.37 & $+3.09$ \\
    \cmdr{}   &  4 & 69.51 & 67.63 & $+1.88$ \\ \bottomrule
  \end{tabular}
\end{table}

GQA requires a single key partition for each query-head group. To quantify the cost of this constraint, \Cref{tab:gqa} compares our shared partition, constructed from the averaged covariance $\Sbar$ in \Cref{eq:avg-graph}, with an otherwise identical per-head construction. We fix the covariance estimates in both cases, isolating the effect of partition sharing.

However, this per-head construction provides shouldn't be viewed as an alternative. It reads $2.2$--$4.8\times$ more keys per group and stores $\gqa$ times as many centroids. The accuracy gap is $4.14$, $3.09$, and $1.88$ points for $\gqa=16$, $8$, and $4$, respectively. Thus, sharing partitions incurs a modest accuracy cost that decreases as fewer query heads share the same keys.

\subsection{Ablations of Gate Strength, Form, and Partition Metric}
\label{app:gate-form}
\begin{figure}[ht]
    \centering
    \includegraphics[width=\linewidth]{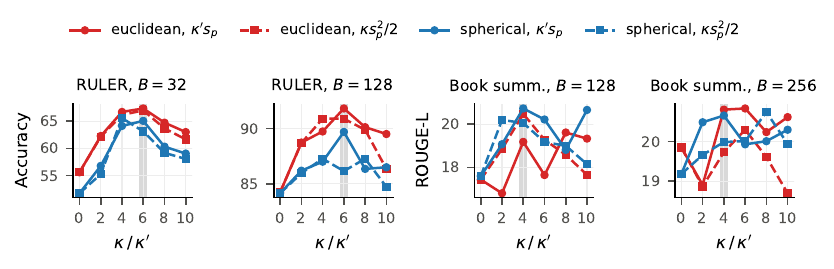}
\caption{Spread-gate performance as a function of gate coefficient on a balanced subset of \textsc{Ruler}-16K and \textsc{InfiniteBench} Book Summarization, at two KV-cache budgets per benchmark. Line style denotes gate form and color denotes partition metric. Solid curves use the linear bias $\kappa'\spread{p}$; dashed curves use the quadratic bias $\kappa\spread{p}^2/2$. Both coefficients take the values shown on the horizontal axis, with zero disabling the gate. Grey bands mark the benchmark-specific linear coefficients used in \Cref{tab:configs}.}

    \label{fig:gate-kappa}
\end{figure}

\paragraph{Ablation Analysis.}
The spread correction involves two design choices: the functional form of the bias and its strength. \Cref{eq:spread-gate} derives a quadratic bias proportional to $\spread{p}^2/2$, while our deployed gate uses a linear bias $\kappa'\spread{p}$. We also compare partitioning by covariance and correlation, using Euclidean and spherical $k$-means, respectively, following \Cref{methods:cov_vs_cor}. \Cref{fig:gate-kappa} evaluates these choices across gate strengths on a retrieval benchmark (a balanced subset of \textsc{Ruler}-16K) and a natural-prose benchmark (a subset of \textsc{InfiniteBench} Book Summarization). \Cref{tab:gate-kappa} averages the four configurations to isolate the effect of gate strength.

\begin{table}[h]
  \caption{Mean score over the four (partition metric, gate form) methods of \Cref{fig:gate-kappa},
  at each coefficient. Averaging over the methods removes the crossings visible in the individual
  curves and shows the effect of the swept coefficient. Best per row in bold.}
  \vspace{0.5em}
  \label{tab:gate-kappa}
  \centering
  \small
  \begin{tabular}{@{}lcccccc@{}}
    \toprule
    Benchmark & $0$ & $2$ & $4$ & $6$ & $8$ & $10$ \\ \midrule
    \textsc{ruler}                & 68.94 & 73.26 & 77.17 & \textbf{77.62} & 75.19 & 73.61 \\
    \textsc{InfiniteBench} summ.  & 18.52 & 19.10 & \textbf{20.20} & 19.67 & 19.58 & 19.42 \\
    \bottomrule
  \end{tabular}
\end{table}

\paragraph{Gate strength.}
Performance follows an inverted-U trend, with the preferred coefficient depending on the workload. Relative to no correction, the best coefficient improves performance by $8.7$ points on \textsc{Ruler} and $1.7$ points on Book Summarization, averaged over the four configurations. The larger variation on \textsc{Ruler} indicates that the spread correction is more consequential for synthetic retrieval than for natural prose. On \textsc{Ruler}, a coefficient of $6$ outperforms $4$ in $6$ of $8$ methods, whereas Book Summarization favors $4$ in $5$ of $8$ methods; the difference is approximately $0.5$ points in either direction. We therefore use $\kappa'=6$ for retrieval (\textsc{Ruler}) and $\kappa'=4$ for natural prose (\textsc{InfiniteBench}) in \Cref{tab:configs}.
This task dependence mirrors that of the query-mix weight (\Cref{app:task-sweeps}).

\paragraph{Gate form.}
The linear and quadratic forms perform similarly at their respective per-cell optima, with the linear form achieving the higher score in $5$ of $8$ cells. Neither form therefore consistently dominates over the range of spreads encountered in our experiments. The distinction is more apparent without per-cell retuning: increasing the swept coefficient to $10$ causes a smaller performance drop for the linear form in $7$ of $8$ cells, by $1.3$ points on average relative to each form's optimum. Since the coefficient is fixed offline and reused across requests and models, this greater robustness motivates our use of the linear form.

\paragraph{Partition metric.}
Logit covariance-based partitioning outperforms logit correlation-based partitioning on \textsc{Ruler} in all $24$ combinations of budget, gate form, and coefficient, by $3.14$ points on average. On Book Summarization, the two metrics differ by less than $0.5$ points on average, with spherical clustering slightly ahead in more cells. This difference is consistent with the role of key magnitude: spherical $k$-means normalizes the transformed keys and removes magnitude information, whereas Euclidean clustering retains it. Magnitude is more informative when retrieval depends on a small number of distinctive keys and less informative when keys are more uniformly distributed, as in natural prose. We therefore use covariance-based partitioning throughout.

\subsection{The Covariance and the Gate, Crossed}
\label{app:sigma-gate}



Here we disentangle the roles of partition and gate geometry. In \Cref{tab:sigma-gate-3x3}, we cross three partition choices with three gate choices: a contiguous positional partition, Euclidean $k$-means under $\Sq=\mI$, or Euclidean $k$-means under the query mix of \Cref{eq:question-mix} with $\mixw=0.5$; and no gate, a gate with $\kappa'=6$ using plain distance, or the corresponding $\Sq$-weighted spread. All differences are paired by sequence and tested with a sign test.



\begin{table}[ht]
\caption{Crossing partition geometry and spread correction on \textsc{Ruler}-16K across KV-cache budgets. Rows vary the partition: a contiguous positional partition, Euclidean $k$-means under $\Sq=\mI$, or Euclidean $k$-means under the query mix of \Cref{eq:question-mix} with $\mixw=0.5$. Columns vary the gate: none, $\kappa'=6$ with spread measured by plain distance, or with spread measured under $\Sq$. Results are averaged over $650$ sequences ($50$ per task); full attention scores $95.08$.}
  \vspace{0.5em}
  \label{tab:sigma-gate-3x3}
  \centering
  \small
  \setlength{\tabcolsep}{4pt}
  \resizebox{\textwidth}{!}{%
  \begin{tabular}{@{}lccccccccccccccc@{}}
    \toprule
    & \multicolumn{3}{c}{$B = 16$}
    & \multicolumn{3}{c}{$B = 32$}
    & \multicolumn{3}{c}{$B = 64$}
    & \multicolumn{3}{c}{$B = 128$}
    & \multicolumn{3}{c}{$B = 256$} \\
    \cmidrule(lr){2-4}
    \cmidrule(lr){5-7}
    \cmidrule(lr){8-10}
    \cmidrule(lr){11-13}
    \cmidrule(lr){14-16}
    & no gate & $\mI$ & $\Sq$
    & no gate & $\mI$ & $\Sq$
    & no gate & $\mI$ & $\Sq$
    & no gate & $\mI$ & $\Sq$
    & no gate & $\mI$ & $\Sq$ \\ \midrule
    positional
      & 19.70 & 20.34 & 20.97
      & 25.32 & 26.91 & 33.65
      & 40.99 & 44.75 & 57.04
      & 51.78 & 57.08 & 67.58
      & 63.34 & 68.23 & 76.76 \\

    $\Sq = \mI$
      & 36.44 & 41.85 & \textbf{45.21}
      & 47.12 & 59.33 & 65.62
      & 61.76 & 74.52 & 77.71
      & 80.94 & 86.59 & 87.36
      & 91.06 & 93.34 & 93.90 \\

    $\Sq$, $\mixw = 0.5$
      & 35.42 & 41.63 & 43.69
      & 50.66 & 62.38 & \textbf{66.14}
      & 69.53 & 79.08 & \textbf{81.37}
      & 84.76 & 89.72 & \textbf{90.48}
      & 93.11 & \textbf{94.27} & 94.18 \\
    \bottomrule
  \end{tabular}
  }
\end{table}




\paragraph{Partitioning and gating both help.}
Replacing the contiguous partition with $k$-means improves performance by $22.7$--$32.5$ points across all budgets. Within the $k$-means partitions, adding the spread correction provides a further improvement of up to $15.5$ points. Thus, both partitioning and gating contribute substantially to retrieval quality.

\paragraph{$\Sq$ improves both components.}
Using $\Sq$ for either partitioning or spread correction generally improves performance over $\Sq=\mI$, with the strongest results obtained when $\Sq$ is used for both. Using $\Sq$ consistently improves performance by up to $6.9$ points over $\Sq=\mI$ in both components, with significant gains from $B=32$ onward.

\paragraph{The two effects overlap.}
Using $\Sq$ in either the partition or the gate recovers much of the gain from using it in both. Thus, the benefits of query-induced geometry in partitioning and gating are complementary but not directly additive: applying $\Sq$ to both remains strongest, but the two improvements provide overlapping ranking information. At $B=256$, all effects beyond partitioning are below one point.

\subsection{The Query Mix Across Tasks}
\label{app:task-sweeps}

We extend the two-workload analysis in \Cref{fig:metric-mismatch-both} to six workloads, using the same normalization: each metric is reported as a fraction of the task's range from $\Sq=\mI$ to its reference. We use ten \textsc{Ruler} sequences per task and ten documents per \textsc{InfiniteBench} task, and additionally show a held-out $\Sq$ for each task. The two question-answering tasks are excluded because their answers are only one to six tokens long, providing too few decode steps to estimate correlations reliably; this leaves $110$ sequences.

\begin{figure}[h]
    \centering
    \includegraphics[width=\linewidth]{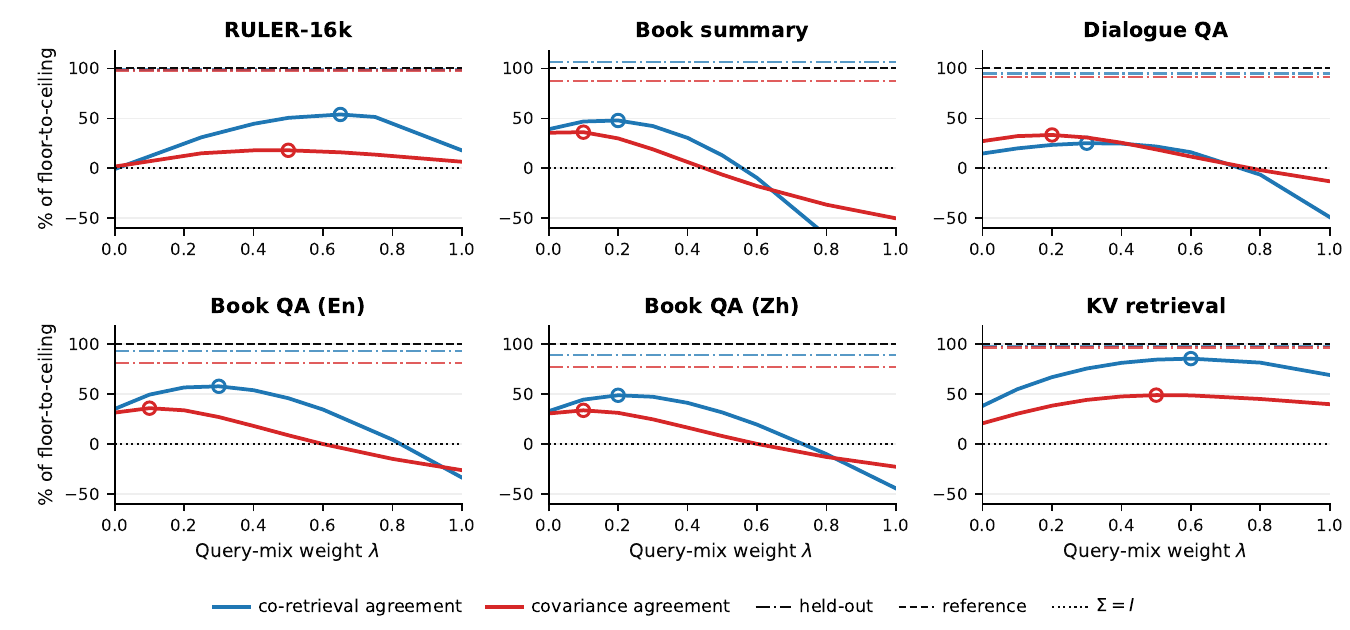}

    \caption{
    Co-retrieval and covariance agreement across query-mix weights.
    Both metrics are normalized to the range from $\Sq=\mI$ to each task's reference on \museg{}, with circles marking their optima. 
    The optimal $\mixw$ tracks context composition rather than length: synthetic retrieval tasks peak near $\mixw=0.6$, while natural-text tasks peak near $\mixw=0.2$. 
    This holds even when \textsc{kv retrieval} uses $124$k tokens, matching the book tasks in length and \textsc{ruler} in composition. Held-out covariances are computed offline from rows excluded from evaluation.}
    
    \label{fig:task-sweeps}
\end{figure}

The optimal mix is task-dependent and tracks the content of the context rather than its length. The two synthetic retrieval tasks peak near $\mixw=0.6$, while the four natural-text tasks peak near $\mixw=0.2$. \textsc{KV Retrieval} helps separate context length from content: despite reaching $124$k tokens, it follows the synthetic \textsc{Ruler} tasks rather than the natural-text tasks. Thus, the preferred mix differs systematically between synthetic retrieval and natural-prose workloads, motivating benchmark-specific mix weights in \Cref{tab:configs}.







\section{Claw-Eval Budget and Repartitioning}
\label{app:claweval-budget}
We report the full repartitioning comparison and results for AgentKV~\citep{agentkv}, an eviction method evaluated at budgets $512$, $1024$, and $1536$.


\begin{figure}[h]
    \centering
    \includegraphics[width=\linewidth]{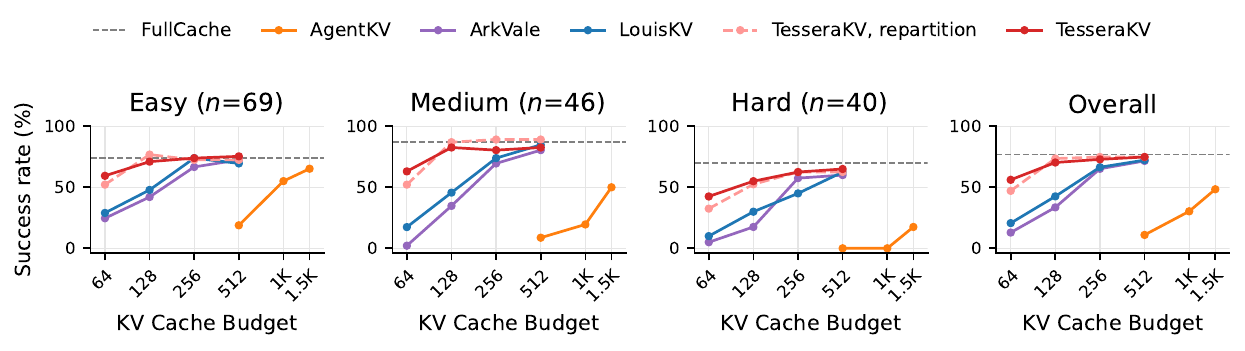}
    \caption{\textsc{Claw-Eval} success rate against the key budget, by difficulty tier and
    task-weighted over all $155$ tasks. The dashed line is full attention. \emph{AgentKV} was run at
    $\budget \in \{512, 1024, 1536\}$ only and the remaining methods at $\budget \le 512$. }
    \label{fig:claweval-budget}
\end{figure}

\begin{table}[h]
  \caption{\textsc{Claw-Eval} success rates (\%) for incremental partitioning and full repartitioning at each turn. Both variants use the same KV budget $\budget$ and evaluation protocol as \cref{tab:claweval}.}
  \label{tab:claweval-repartition}
  \centering
  \small
  \resizebox{\textwidth}{!}{%
    \begin{tabular}{@{}lcccc|cccc|cccc|cccc@{}}
      \toprule
      & \multicolumn{4}{c}{Easy} & \multicolumn{4}{c}{Medium} & \multicolumn{4}{c}{Hard} & \multicolumn{4}{c}{Overall} \\
      \cmidrule(lr){2-5} \cmidrule(lr){6-9} \cmidrule(lr){10-13} \cmidrule(l){14-17}
      KV budget $\budget$ & $64$ & $128$ & $256$ & $512$ & $64$ & $128$ & $256$ & $512$ & $64$ & $128$ & $256$ & $512$ & $64$ & $128$ & $256$ & $512$ \\ \midrule
      $\text{\mymethod{}}_{\mathrm{Repartition}}$ & 52.17 & 76.81 & 72.46 & 72.46 & 52.17 & 86.96 & 89.13 & 89.13 & 32.50 & 52.50 & 62.50 & 62.50 & 47.10 & 73.55 & 74.84 & 74.84 \\
      $\text{\mymethod{}}$ & 59.42 & 71.01 & 73.91 & 75.36 & 63.04 & 82.61 & 80.43 & 82.61 & 42.50 & 55.00 & 62.50 & 65.00 & 56.13 & 70.32 & 72.90 & 74.84 \\
      \bottomrule
    \end{tabular}%
  }
\end{table}

\Cref{tab:claweval-repartition} compares incremental partitioning with rebuilding the partition over the full context at every turn. In overall success rate, incremental partitioning leads by $9.03$ points at $\budget=64$, trails by $3.23$ and $1.94$ points at $\budget=128$ and $256$, and ties at $\budget=512$.

In \Cref{fig:claweval-budget}, the retrieval methods differ most at tight budgets.
At $\budget = 64$, \mymethod{} reaches $56.1\%$, compared with $20.7\%$ for LouisKV and $12.9\%$ for ArkVale. At $\budget = 512$, their success rates rise to $74.8\%$, $72.3\%$, and $71.6\%$, respectively, against $76.8\%$ for full attention.

Eviction behaves differently in kind:
AgentKV reaches $30.3\%$ at $\budget = 1024$, below \mymethod{}'s $56.1\%$ at one-sixteenth of that budget, and solves none of the hard tier until $\budget = 1536$, three times the largest retrieval budget tested. \textsc{Claw-Eval} carries the
cache across turns, so a key discarded in one turn is unavailable in the next, whereas the retrieval
methods retain the cache and re-select from it at every step.

\section{Offloading the KV Cache to CPU Memory}
\label{app:offload}
When the KV cache does not fit in device memory, it can be kept in host memory and fetched on demand. Our setup follows RetroInfer~\citep{chen2025retroinfer}: the full KV cache resides in CPU memory, and a GPU-resident wave buffer caches recently retrieved keys, so that each decoding step transfers only the selected keys that miss the buffer. All measurements in this section use a single B200 GPU connected to host memory over PCIe~5.

\subsection{Comparison with RetroInfer}
\label{app:retroinfer}
RetroInfer retrieves a subset of clusters and estimates the contributions of those not retrieved, whereas \mymethod{} excludes non-retrieved pages from the attention output. These approaches are orthogonal: \mymethod{} can readily incorporate RetroInfer's estimation mechanism. We therefore evaluate \mymethod{} both with and without this mechanism. We use the released RetroInfer implementation and evaluate both methods on \textsc{Ruler}-16K with \cmdr{}, using 64-key pages and clusters to ensure that both methods read the same number of keys at each budget.

\paragraph{GPU-resident KV cache.}
\Cref{tab:retroinfer-gpu} keeps the KV cache in device memory. \mymethod{} attains higher accuracy at the smaller budget ($79.97$ versus $66.23$ at $128$ keys, and $85.01$ with RetroInfer's estimation). Increasing the retrieval budget from $128$ to $512$ keys has little effect on decoding latency for either method, owing to the high bandwidth of HBM: at batch size $64$, the per-step latency changes by about $1\%$ for \mymethod{} and $4\%$ for RetroInfer. The two methods therefore have comparable decoding latency at both budgets, with \mymethod{} $3$--$6\%$ faster per step at batch size $64$.

\begin{table}[h]
  \caption{\textsc{Ruler}-16K accuracy and decoding latency with the KV cache resident in GPU memory, for
  \cmdr{} on one B200. Pages and clusters both hold $64$ keys, so RetroInfer with $2$ and $8$ clusters
  reads the same number of keys as \mymethod{} at $\budget=128$ and $512$. ``+ est.'' adds RetroInfer's
  estimation of the non-retrieved clusters. Latency is per attention layer at batch sizes $1$ and $4$,
  and per decoding step at batch sizes $32$ and $64$.}
  \vspace{0.5em}
  \label{tab:retroinfer-gpu}
  \centering
  \small
  \resizebox{\textwidth}{!}{%
  \begin{tabular}{@{}lcccc@{}}
    \toprule
    Method & \textsc{Ruler}-16K & $\mu$s/layer, batch $1$ / $4$ & ms/step, batch $32$ & ms/step, batch $64$ \\ \midrule
    RetroInfer, $2$ clusters (+ est.)              & $66.23$          & $24.7$ / $28.5$ & $8.69$ & $12.49$ \\
    \mymethod{}, $\budget=128$                      & $79.97$          & $21.2$ / $24.2$ & $8.45$ & $12.06$ \\
    \mymethod{}, $\budget=128$ + RetroInfer est.    & $\mathbf{85.01}$ & $24.3$ / $28.1$ & $8.60$ & $12.32$ \\ \midrule
    RetroInfer, $8$ clusters (+ est.)              & $86.50$          & $25.7$ / $33.5$ & $8.95$ & $13.01$ \\
    \mymethod{}, $\budget=512$                      & $86.46$          & $21.6$ / $27.0$ & $8.41$ & $12.19$ \\
    \mymethod{}, $\budget=512$ + RetroInfer est.    & $\mathbf{86.60}$ & $24.7$ / $31.0$ & $8.69$ & $12.47$ \\
    \bottomrule
  \end{tabular}
  }
\end{table}

\paragraph{KV cache offloaded to CPU memory.}
With the KV cache in host memory, retrieval becomes sensitive to PCIe bandwidth. \Cref{tab:retroinfer-offload} uses RetroInfer's wave buffer at capacities of $800$ keys ($5\%$ of the context) and $1{,}536$ keys per layer and head. With the $800$-key buffer, \mymethod{} at $\budget=128$ with RetroInfer's estimation comes within $1.49$ points of RetroInfer at $512$ keys ($85.01$ versus $86.50$). Because \mymethod{} selects $4\times$ fewer keys, the same buffer is relatively larger for it and yields a higher hit rate ($0.96$ versus $0.73$); together with the smaller budget, this reduces PCIe traffic by $20\times$ and decoding latency by $37\%$ at batch size $64$ ($13.13$ versus $20.91$\,ms).

\begin{table}[h]
  \caption{\textsc{Ruler}-16K accuracy and decoding latency with the KV cache offloaded to CPU memory, for
  \cmdr{} on one B200 over PCIe~5. Pages and clusters both hold $64$ keys, so RetroInfer with $2$ and $8$
  clusters reads the same number of keys as \mymethod{} at $\budget=128$ and $512$. ``+ est.'' adds
  RetroInfer's estimation of the non-retrieved clusters. Latency is per decoding step at a wave-buffer
  capacity of $800$ and $1{,}536$ keys per layer and head.}
  \vspace{0.5em}
  \label{tab:retroinfer-offload}
  \centering
  \small
  \begin{tabular}{@{}lccc@{}}
    \toprule
    Method & \textsc{Ruler}-16K & ms/step, batch $32$ & ms/step, batch $64$ \\
           &                    & ($800 \to 1{,}536$) & ($800 \to 1{,}536$) \\ \midrule
    RetroInfer, $2$ clusters (+ est.)              & $66.23$          & $9.19 \to 9.30$   & $13.15 \to 13.23$ \\
    \mymethod{}, $\budget=128$                      & $79.97$          & $9.04 \to 9.14$   & $12.92 \to 12.88$ \\
    \mymethod{}, $\budget=128$ + RetroInfer est.    & $\mathbf{85.01}$ & $9.15 \to 9.17$   & $13.13 \to 13.10$ \\ \midrule
    RetroInfer, $8$ clusters (+ est.)              & $86.50$          & $13.05 \to 10.30$ & $20.91 \to 15.46$ \\
    \mymethod{}, $\budget=512$                      & $86.46$          & $11.41 \to 9.82$  & $17.85 \to 14.75$ \\
    \mymethod{}, $\budget=512$ + RetroInfer est.    & $\mathbf{86.60}$ & $11.62 \to 10.10$ & $18.03 \to 14.95$ \\
    \bottomrule
  \end{tabular}
\end{table}

\subsection{End-to-End Cost on \textsc{Claw-Eval}}
\label{app:claweval-cost}
\mymethod{} at $\budget=128$ comes close to the success rate of LouisKV at $\budget=512$ ($70.32$ versus $72.26$, \Cref{tab:claweval}). We compare the cost of completing \textsc{Claw-Eval} tasks at these two budgets. This task has on average $12$ turns, each with about $10$K input tokens and $279$ output tokens.

\paragraph{Time to first token.}
\Cref{tab:claweval-ttft} reports TTFT at batch size $1$, including index construction and maintenance. Summed over the turns of a task, TTFT is $5.4$\,s for \mymethod{} and $6.7$\,s for LouisKV. Index construction and maintenance account for $1.85$\,s and $2.65$\,s, respectively. The two methods add a similar overhead on the median task. LouisKV rebuilds its index at every turn, which becomes increasingly costly in longer sessions and raises its average TTFT by $1.2$\,s over \mymethod{}.

\begin{table}[h]
  \caption{Time to first token on \textsc{Claw-Eval} at batch size $1$, summed over the turns of a task,
  with its breakdown into index construction and maintenance, and pre-fill. KV-cache write-back
  to CPU memory is included for the offloaded configurations.}
  \vspace{0.5em}
  \label{tab:claweval-ttft}
  \centering
  \small
  \begin{tabular}{@{}lccc@{}}
    \toprule
    Method & TTFT per task & Index build & Pre-fill \\ \midrule
    \mymethod{}, $\budget=128$     & $5.44$\,s & $1.85$\,s & $3.59$\,s \\
    LouisKV, $\budget=512$ & $6.68$\,s & $2.65$\,s & $4.03$\,s \\
    \bottomrule
  \end{tabular}
\end{table}

\begin{figure}[t]
    \centering
    \includegraphics[width=\linewidth]{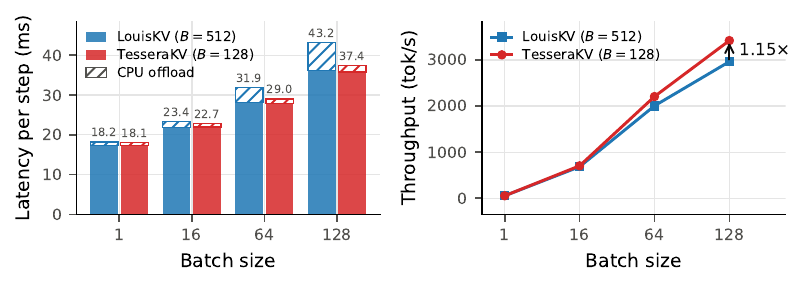}
    \vspace{-1.5em}
    \caption{Decoding cost on \textsc{Claw-Eval} with the KV cache offloaded to CPU memory, for \mymethod{} at
    $\budget=128$ and LouisKV at $\budget=512$, which reach similar
    success rates (\Cref{tab:claweval}). Both use a RetroInfer-style wave buffer with the same $768$-key GPU
    cache on one B200 over PCIe~5. \emph{Left:} latency per decoding step, split into GPU compute and
    KV transfer from host memory. \emph{Right:} decoding throughput.}
    \label{fig:decode-batch}
\end{figure}

\paragraph{Decoding with the KV cache offloaded to CPU memory.}
We next evaluate KV cache offloading to host memory, fetching selected keys on demand using the same wave-buffer design for both methods. Each method uses an identical, fixed-size GPU wave buffer holding $768$ keys. 
At $\budget=128$, \mymethod{} selects $4\times$ fewer keys per step than LouisKV at $\budget=512$. With the same cache capacity, this smaller working set allows a larger fraction of \mymethod{}'s selected pages to remain resident in the GPU cache, resulting in a higher page hit rate ($82\%$ vs. $66\%$). Together, these factors reduce KV traffic over PCIe by $6.1\times$ ($0.38$ vs. $2.31$\,MiB per sequence per step). As batch size increases, KV transfers account for a growing share of step time, further amplifying this advantage (\cref{fig:decode-batch}). \mymethod{} achieves $10\%$ higher throughput at batch size $64$ and $15\%$ higher throughput at batch size $128$ ($3{,}418$ vs. $2{,}962$ tokens/s), while reducing per-step latency by $13\%$.

\section{Additional Inference Details}

\subsection{Inference Implementation}
  \label{app:inference-impl}
  Our SGLang implementation combines asynchronous index construction during prefill with GPU-resident page selection
  and sparse attention during decode. KV paging is applied to global-attention layers, while sliding-window layers
  retain their local attention pattern. The full KV cache remains resident on the GPU; the index stores page
  summaries and physical token addresses.

  \paragraph{Prefill and partition construction.}
  Both our implementation and the dense baseline use FlashAttention~4 for prefill. We accumulate query statistics in
  FP32 across prefill chunks without retaining the full query tensor. These statistics determine the covariance-
  weighted key transformation used for clustering. To accelerate Lloyd iterations, we adapt the tiled Triton
  assignment kernel from \texttt{flash-kmeans}, using TF32 matrix multiplication with FP32 accumulation while
  retaining FP32 features and centroids. Centroid updates and the final spread correction follow our reference
  implementation.
  Once a layer's inputs are available, a background host worker dispatches index construction onto a separate CUDA
  stream, allowing it to overlap with subsequent layers' forward computation. CUDA events enforce data dependencies,
  and indexed decode waits for construction to complete. Consequently, overlap can hide part of the indexing cost;
  any unfinished work remains on the critical path.

  \paragraph{Page scoring.}
  The decode indexer computes query--centroid scores using batched matrix multiplication with BF16 inputs and
  outputs. Triton kernels then apply the spread correction, compute per-query-head softmax in FP32, average
  probabilities within each GQA group, and select pages until the key budget is filled. The optimized selection path
  extracts candidates in top-$k$ blocks rather than fully sorting all pages, truncating the final selected page when
  necessary. It packs physical token addresses without copying the selected KV entries. These operations are fused
  where supported; larger page counts use separate softmax and selection kernels.

  \paragraph{Sparse attention.}
  A custom Triton kernel reads the selected keys and values directly from their physical locations in the KV cache,
  together with the always-attended sink tokens and question/generated tail. It partitions the attention computation
  across 16 splits, maintains FP32 online-softmax statistics and accumulators, and combines the partial outputs with
  a subsequent merge kernel. This avoids materializing a compact KV tensor and reduces decode-time KV reads. The
  dense baseline uses the FlashAttention~4 backend.






\subsection{Latency Across Context Lengths and Batch Sizes}
\label{app:latency}

\begin{figure}[h]
    \centering
    \includegraphics[width=0.9\linewidth]{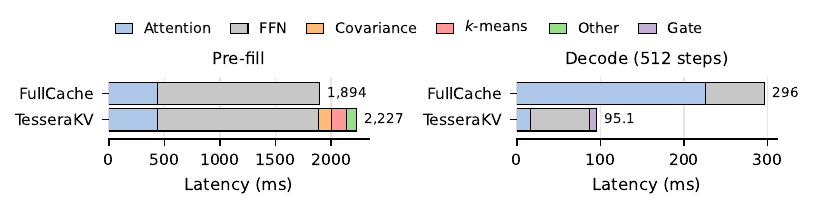}\\
    \includegraphics[width=0.9\linewidth]{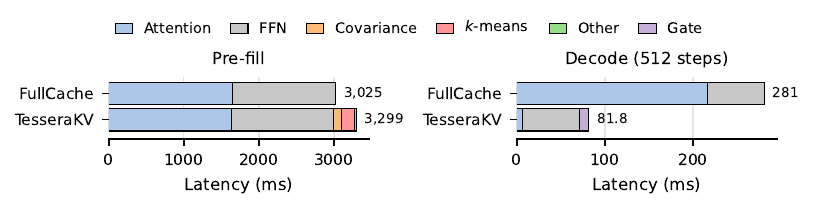}
    \caption{Profiled latency breakdown for one layer with global attention at $32$k context and batch size $64$ (top) and $128$k context and batch size $16$ (bottom), with $\budget=512$. Bars sum component times without asynchronous overlap; decode totals cover $512$ steps.}
    \label{fig:latency-cells}
\end{figure}

\Cref{fig:latency-cells} extends the breakdown in \cref{fig:latency} to $32$k context at batch size $64$ and $128$k at batch size $16$. All three configurations use $512$ decode steps.

Prefill attention becomes more expensive at longer contexts, while the feed-forward component changes much less across these configurations.

The added costs remain similar across configurations: covariance estimation takes $111$, $106$, and $103$\,ms, while index construction takes $227$, $172$, and $199$\,ms at $32$k, $64$k, and $128$k, respectively. When profiled component times are summed without overlap, the relative increase in prefill time falls from $17.6\%$ to $11.5\%$ and $9.1\%$ as global attention becomes more expensive.

During decoding, the indexer accounts for an increasing share of the profiled paging cost: $33\%$, $48\%$, and $61\%$ at $32$k, $64$k, and $128$k, respectively. More pages must be scored as the context grows, while the attention budget $\budget$ remains fixed. Sparse attention alone is $13.6\times$ to $33.2\times$ faster than full attention, but the gain is $8.7\times$ to $12.4\times$ when page selection is included in the summed component times. Page selection therefore limits the projected decode gain at longer contexts.

These per-layer measurements include the feed-forward block; \cref{fig:throughput-batch} reports whole-model throughput.

\section{Validating the Modelling Approximations}
\label{app:approximations}
\begin{figure}[h]
    \centering
    \includegraphics[width=\linewidth]{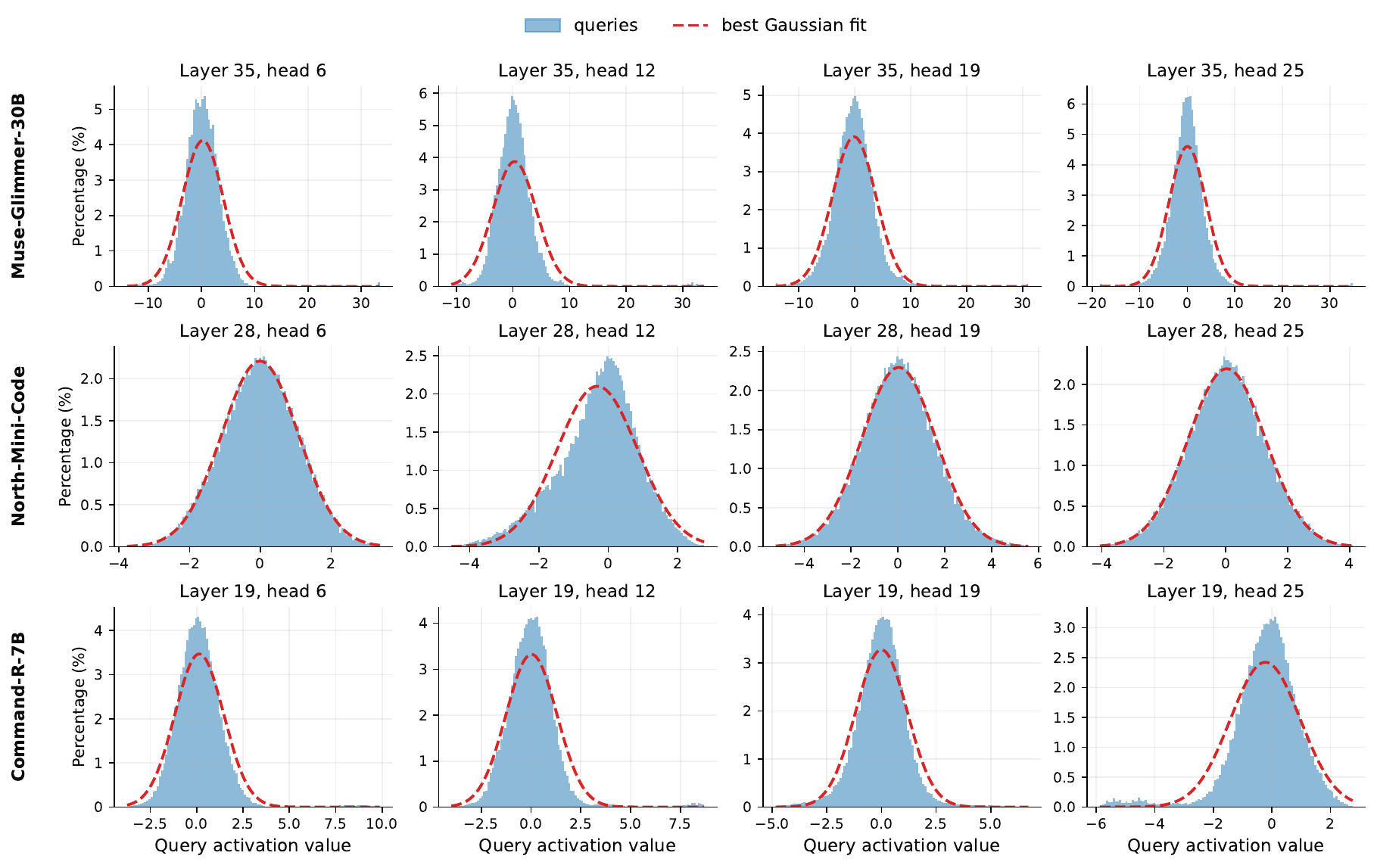}
    \vspace{-1.5em}
    \caption{Distribution of query activations for four query heads of one paged layer of each model (layers $35$, $28$ and $19$),
    with the best Gaussian fit overlaid (dashed). Each panel pools all dimensions of the last $2{,}048$
    pre-fill queries of a \textsc{Ruler}-16K document.}
    \label{fig:app:query-gauss}
\end{figure}

The partition and gate rely on three modelling approximations, which we validate on the evaluated models.
\paragraph{Query distribution.}
Following prior work that models LLM hidden states as approximately Gaussian~\citep{training-free-activation-sparsity,ea}, 
we model decoding queries for each head as samples from a distribution with covariance $\Sq$. 
\Cref{fig:app:query-gauss} plots the query activation distributions for individual heads across the evaluated models, alongside their best Gaussian fits. 
Most heads exhibit unimodal, centred, bell-shaped distributions that closely follow the Gaussian fits, although some are more sharply peaked and a few are heavy-tailed.

\begin{figure}[t]
    \centering
    \includegraphics[width=\linewidth]{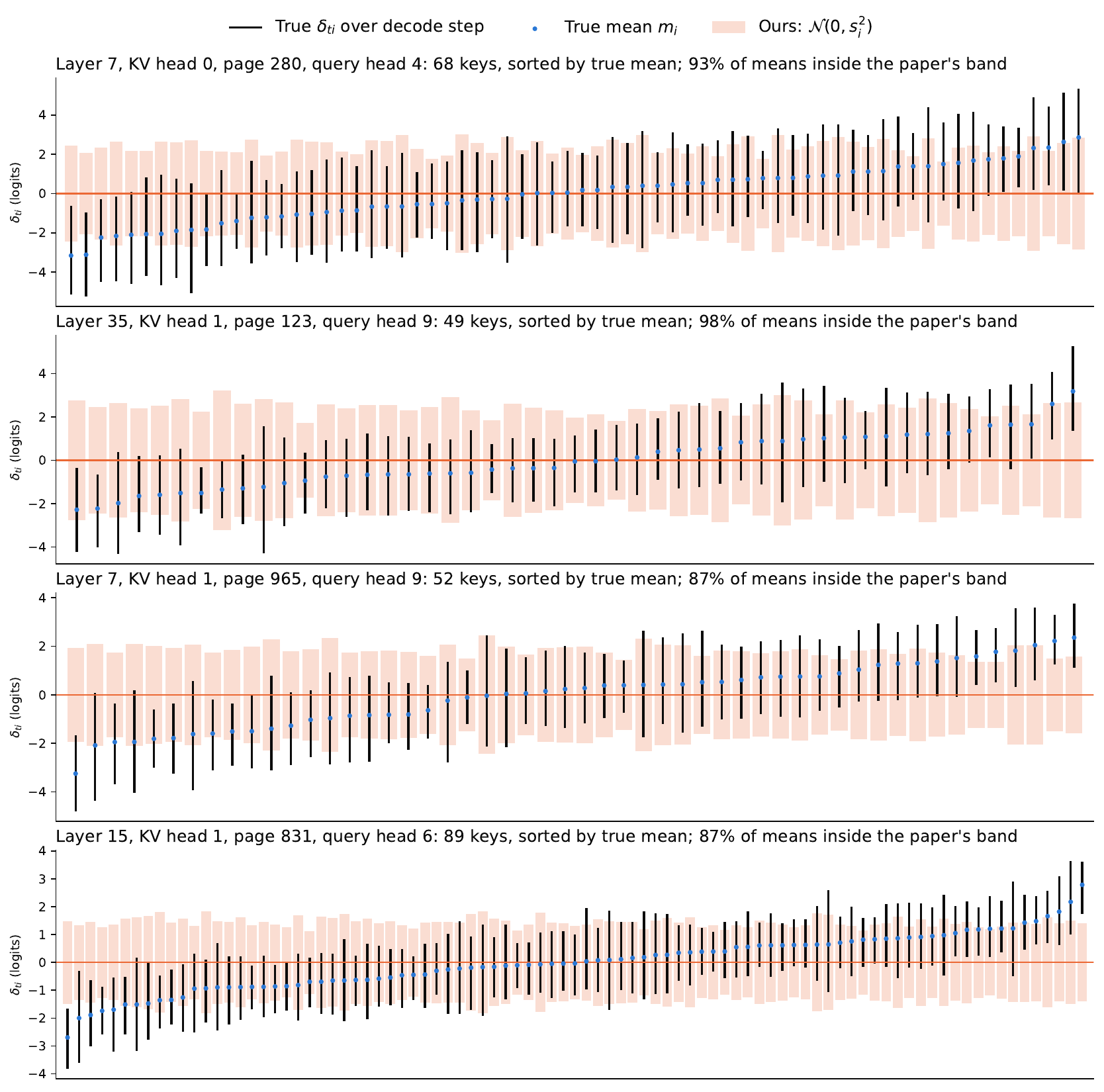}
    \vspace{-1.5em}
    \caption{Per-key deviations $\delta_{ti}$ against the model $\mathcal{N}(0,\spread{i}^2)$ for every key
    of four example pages from different layers and heads of \museg{} on \textsc{InfiniteBench} Book
    Summarization. Keys are sorted by their mean. Black bars show the empirical $10$--$90\%$ range of
    $\delta_{ti}$ over $96$ decode queries and blue dots its mean; shaded bands show the $10$--$90\%$
    range of $\mathcal{N}(0,\spread{i}^2)$, with $\spread{i}$ estimated before decoding.}
    \label{fig:app:delta-page-keys}
\end{figure}

\begin{figure}[t]
    \centering
    \includegraphics[width=\linewidth]{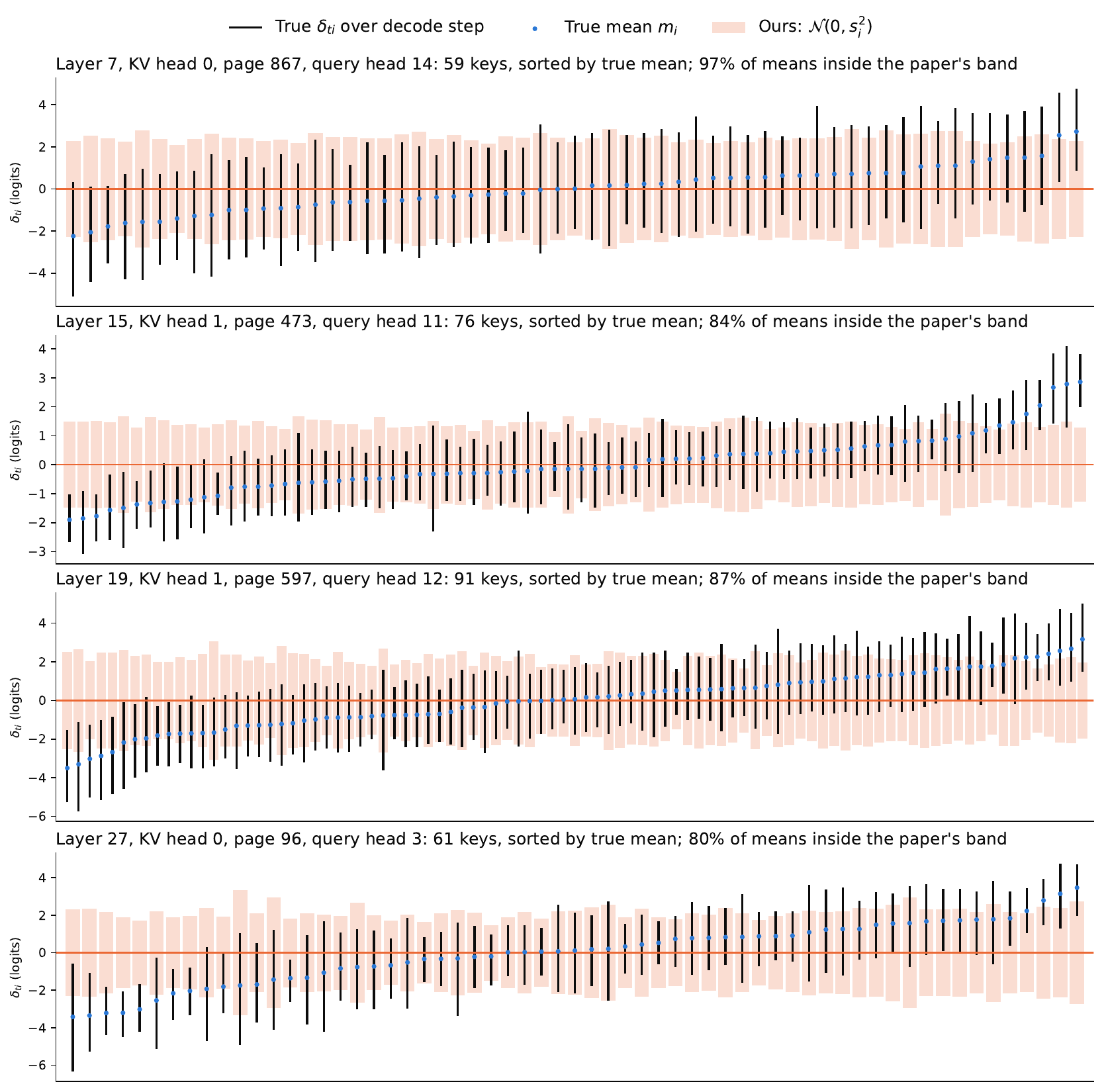}
    \vspace{-1.5em}
    \caption{Four further example pages, as in \cref{fig:app:delta-page-keys}, drawn from four other layers
    of \museg{} on \textsc{InfiniteBench} Book Summarization.}
    \label{fig:app:delta-page-keys-2}
\end{figure}

\paragraph{Gaussian model for per-key logit deviations.} The gate models each key's deviation from its page score, $\delta_{ti}$, as $\mathcal{N}(0,\spread{i}^2)$ across decoding queries. 
\Cref{fig:app:delta-page-keys,fig:app:delta-page-keys-2} compare this model with the empirical distribution for keys sampled across different layers and heads in Book summarization tasks.
The predicted spread $\spread{i}$, estimated before decoding, closely matches the observed per-key variation (within a factor of $1.2$--$1.4$). 
Individual key means are not exactly zero. However, they fall largely within the predicted $10$--$90\%$ range from the zero mean gaussian model.
After centering each key's offsets by its empirical mean, its standard deviation equal to $0.99$, supporting the Gaussian shape assumption. 

\paragraph{Approximating per-key spreads with page means.}
\Cref{eq:spread} approximates each key's spread $\spread{i}$ using its page mean $\spread{p}$. 
This is justified by the low variation in spreads within each page, as the partition groups keys with similar transformed offsets. 
The median within-page coefficient of variation is $0.14$ on Book Summarization and $0.16$ on \textsc{Ruler}-16K, 
consistent with the nearly uniform band heights in \cref{fig:app:delta-page-keys}. 
This approximation changes the estimated page correction by only $2.7\%$ and $2.3\%$ at the median respectively.

\section{Limitations}
\label{app:limitations}
Our method is designed for NoPE global full-attention layers and does not directly support rotary positional embeddings (RoPE). Although NoPE is increasingly used in the global attention layers of hybrid long-context models, some models retain RoPE. 
Here we empirically evaluate TesseraKV on two RoPE-based models, Llama-3.1-8B and Ministral-3-8B in \Cref{tab:rope_transfer}. We consider two variants: (i) estimating the query covariance directly from post-RoPE queries, and (ii) combining TesseraKV with Expected Attention (EA)~\citep{ea}, which approximates the position-dependent rotations using a single average rotation matrix. Both variants outperform LouisKV across majority of the evaluated retrieval budgets, with accuracy improvements up to 5.7 points for the first variant and to 5.8 points for the EA-enhanced variant. EA further provides a small but consistent improvement over direct post-RoPE covariance estimation, with larger gains at longer context lengths. These results suggest that TesseraKV can transfer empirically to RoPE-based models, although its theoretical analysis remains limited to NoPE attention. A more comprehensive investigation of RoPE support is left for future work. A more systematic investigation of this approach is left to future work.
\begin{table}[h]
\caption{
Empirical transfer of \mymethod{} to RoPE-based models. We report results on Llama-3.1-8B-Instruct and Ministral-3-8B-Instruct at 8K and 16K context lengths under different KV budgets. FullCache denotes attention over the entire KV cache. Best results among the compressed methods are shown in bold.
}
\label{tab:rope_transfer}
\centering
\small
\resizebox{\textwidth}{!}{%
\begin{tabular}{@{}llccccc|ccccc@{}}
\toprule
\multirow{2}{*}{Model} & \multirow{2}{*}{Method}
& \multicolumn{5}{c}{8K}
& \multicolumn{5}{c}{16K} \\
\cmidrule(lr){3-7} \cmidrule(l){8-12}
& KV budget $\budget$
& 32 & 64 & 128 & 256 & Avg.
& 32 & 64 & 128 & 256 & Avg. \\
\midrule
\multirow{4}{*}{Llama-3.1-8B}
& FullCache
& \multicolumn{5}{c|}{94.06}
& \multicolumn{5}{c}{92.69} \\\cmidrule(l){2-12}
& LouisKV
& 45.98 & 62.63 & 77.70 & 88.28 & 68.65
& 42.17 & 54.96 & 70.44 & 82.58 & 62.54 \\
& \mymethod{}
& \textbf{50.84} & 71.35 & 85.11 & \textbf{90.19} & 74.37
& 40.91 & 59.54 & 75.18 & 86.07 & 65.43 \\
& \mymethod{} + EA
& 50.68 & \textbf{71.53} & \textbf{85.57} & 90.06 & \textbf{74.46}
& \textbf{42.33} & \textbf{61.14} & \textbf{77.54} & \textbf{86.40} & \textbf{66.85} \\
\midrule
\multirow{4}{*}{Ministral-3-8B}
& FullCache
& \multicolumn{5}{c|}{93.95}
& \multicolumn{5}{c}{93.06} \\\cmidrule(l){2-12}
& LouisKV
& 53.12 & 73.88 & \textbf{87.54} & \textbf{92.77} & 76.83
& 46.19 & 69.04 & 82.39 & 88.49 & 71.53 \\
& \mymethod{}
& 57.26 & \textbf{75.27} & 84.92 & 91.16 & 77.15
& 53.09 & 70.03 & 80.80 & 89.52 & 73.36 \\
& \mymethod{} + EA
& \textbf{58.64} & 74.85 & 86.60 & 91.97 & \textbf{78.02}
& \textbf{55.87} & \textbf{72.39} & \textbf{82.66} & \textbf{90.23} & \textbf{75.29} \\
\bottomrule
\end{tabular}%
}
\end{table}

\section{Relation to Other Axes of Efficient Inference}
\label{app:efficiency}

Reducing KV-cache traffic is one axis in the broader effort to reduce inference cost, and these axes are largely complementary. Weight-side methods reduce model size: post-training quantization lowers parameter precision \citep{awq, gptq, Turboquant, KVarN, lqer, qera}, while factorization methods compress weight matrices directly \citep{mha2mla, svd-llm, a3}. Cache-side methods, the focus of this paper, change how cached keys and values are stored or accessed while leaving model weights unchanged.

Our method also naturally extends to multi-agent workloads \citep{chain-of-agents,latentmas,tot,agentic-reasoning, xu2026rethinking}, where multiple tool-using agents maintain long, multi-turn sessions of the kind evaluated in this work. Each agent can independently maintain and page its own KV cache.

On the system side, our implementation keeps the partitioned KV cache in HBM. The same partitioned representation could instead be offloaded to CPU memory, following prior systems for KV-cache offloading \citep{oasisKV, louiskv, chen2025retroinfer}. We leave this extension to future work.



\end{document}